\documentclass{article}

\usepackage[T1]{fontenc}
\usepackage{tgtermes}

\usepackage{amssymb}
\newcommand{\mathbold}[1]{\ensuremath{\boldsymbol{\mathbf{#1}}}}

\usepackage{scalefnt,letltxmacro}
\LetLtxMacro{\oldtextsc}{\textsc}
\renewcommand{\textsc}[1]{\oldtextsc{\scalefont{1.10}#1}}
\usepackage[ttdefault=true]{AnonymousPro}

\usepackage{amsmath}

 \usepackage[
  paper  = letterpaper,
      textwidth=5.5in,
    textheight=9in,
    headheight=12pt,
    headsep=25pt,
    footskip=30pt,
      top    = 1.0in,
   bottom = 1.0in,
   ]{geometry}
   \usepackage[english]{babel}
\usepackage[parfill]{parskip}
\usepackage{afterpage}
\usepackage{enumitem}
\usepackage{framed}
\usepackage{xspace}

\usepackage{lineno}

\usepackage{ragged2e}

\newcounter{parcount}

\DeclareRobustCommand{\parhead}[1]{\textbf{#1}~}
\usepackage[usenames,dvipsnames]{xcolor}
\definecolor{shadecolor}{gray}{0.9}

\newcommand{\red}[1]{\textcolor{BrickRed}{#1}}

\newcommand{\green}[1]{\textcolor{OliveGreen}{#1}}

\usepackage{graphicx}
\usepackage[labelfont=bf]{caption}
\usepackage[format=hang]{subcaption}

\usepackage{booktabs, array}

\usepackage[numbers]{natbib}

\usepackage[algoruled]{algorithm2e}
\usepackage{listings}
\usepackage{fancyvrb}
\fvset{fontsize=\small}

\usepackage[colorlinks,linktoc=all]{hyperref}
\usepackage[all]{hypcap}
\hypersetup{citecolor=MidnightBlue}
\hypersetup{linkcolor=MidnightBlue}
\hypersetup{urlcolor=MidnightBlue}

\usepackage
[acronym,smallcaps,nowarn,section,nogroupskip,nonumberlist]{glossaries}
\glsdisablehyper{}

\lstdefinestyle{mystyle}{
    commentstyle=\color{OliveGreen},
    numberstyle=\tiny\color{black!60},
    stringstyle=\color{BrickRed},
    basicstyle=\ttfamily\scriptsize,
    breakatwhitespace=false,
    breaklines=true,
    captionpos=b,
    keepspaces=true,
    numbers=none,
    numbersep=5pt,
    showspaces=false,
    showstringspaces=false,
    showtabs=false,
    tabsize=2
}
 \newacronym{CUBO}{cubo}{$\chi$ upper bound}
\newacronym{BBVI}{bbvi}{black box variational inference}
\newacronym{ELBO}{elbo}{evidence lower bound}
\newacronym{EP}{ep}{expectation propagation}
\newacronym{GP}{gp}{Gaussian process}
\newacronym{VI}{vi}{Variational inference}
\newacronym{KL}{kl}{Kullback-Leibler}
\newacronym{MCMC}{mcmc}{Markov chain Monte Carlo}
\newacronym{HMC}{hmc}{Hamiltonian Monte Carlo}
\newacronym{chiVI}{chivi}{$\chi$-divergence variational inference}
\newacronym{klVI}{klvi}{$\operatorname{KL}(q || p)$ variational inference}
 
\newcommand{\g}{\,|\,}
\renewcommand{\gg}{\,\|\,}

\newtheorem{theorem}{Theorem}

\newcommand{\mbw}{\mathbold{w}}
\newcommand{\mbx}{\mathbold{x}}

\newcommand{\mbz}{\mathbold{z}}

\newcommand{\mbK}{\mathbold{K}}
\newcommand{\mbL}{\mathbold{L}}

\newcommand{\mblambda}{\mathbold{\lambda}}

\usepackage[numbers, final]{nips_2017}

\usepackage[utf8]{inputenc} \usepackage[T1]{fontenc}    \usepackage{hyperref}       \usepackage{url}            \usepackage{booktabs}       \usepackage{amsfonts}       \usepackage{nicefrac}       \usepackage{microtype}      \usepackage{times}

\title{Variational Inference via \\ $\chi$ Upper Bound Minimization}

\author{
  Adji B. Dieng\\
 Columbia University\\
    \And
  Dustin Tran\\
  Columbia University\\
    \And
  Rajesh Ranganath\\
 Princeton University\\
    \AND
  John Paisley\\
 Columbia University\\
     \And
   David M. Blei\\
 Columbia University\\
  }

\begin{document}
\maketitle

\begin{abstract}
\vskip 0.1in
\gls{VI} is widely used as an efficient alternative to \acrlong{MCMC}. It
posits a family of approximating distributions $q$ and finds the closest member
to the exact posterior $p$. Closeness is usually measured 
via a divergence $D(q || p)$ from $q$ to $p$. 
While successful, this approach also has problems.  
Notably, it typically leads to underestimation of the posterior variance. 
In this paper we propose \glsunset{chiVI}\gls{chiVI}, a 
black-box variational inference algorithm that minimizes 
 $D_{\chi}(p || q)$, the $\chi$-divergence from $p$ to $q$.~\gls{chiVI} minimizes an upper bound
of the model evidence, which we term the \gls{CUBO}.  Minimizing the
\gls{CUBO} leads to improved posterior uncertainty, and it can also be used
with the classical \gls{VI} lower bound (\acrshort{ELBO})
to provide a sandwich estimate of the
model evidence.  We study \gls{chiVI} on three models: probit regression, 
Gaussian process classification, and a Cox process model of basketball plays. 
When compared to \acrlong{EP} and classical \gls{VI}, \gls{chiVI} produces 
better error rates and more accurate estimates of posterior variance.
\end{abstract}

\section{Introduction}
\label{sec:introduction}

Bayesian analysis provides a foundation for reasoning with
probabilistic
models. We first set a joint distribution $p(\mbx, \mbz)$ of latent variables
$\mbz$ and observed variables $\mbx$. We then analyze data through the
posterior, $p(\mbz\g \mbx)$.  In most applications, the posterior is
difficult to compute because the marginal likelihood $p(\mbx)$ is
intractable.  We must use approximate posterior inference methods such
as Monte Carlo~\citep{Robert:2004} and variational
inference~\citep{Jordan:1999}. This paper focuses on variational inference.

Variational inference approximates the posterior using optimization.
The idea is to posit a family of approximating distributions and then
to find the member of the family that is closest to the posterior.
Typically, closeness is defined by the \gls{KL} divergence
$\operatorname{KL}(q \gg p)$, where $q(\mbz; \mblambda)$ is a
variational family indexed by parameters $\mblambda$. This approach,
which we call \glsunset{klVI}\gls{klVI}, also provides the \gls{ELBO},
a convenient lower bound of the model evidence $\log p(\mbx)$.

\gls{klVI} scales well and is suited to
applications that use complex models to analyze large data sets~\citep{hoffman2013stochastic}. But
it has drawbacks. For one, it tends to favor underdispersed
approximations relative to the exact
posterior~\citep{murphy2012machine,bishop2006pattern}. This
produces difficulties with light-tailed posteriors when the variational
distribution has heavier tails. For example, \gls{klVI} for Gaussian
process classification typically uses a Gaussian approximation; this
leads to unstable optimization and a poor
approximation~\citep{hensmantilted}.

One alternative to \glsunset{klVI}\gls{klVI} is \glsreset{EP}\gls{EP},
which enjoys good empirical performance on models with light-tailed
posteriors~\citep{minka2001family,kuss2005assessing}.
Procedurally, \gls{EP} reverses the arguments in the \gls{KL}
divergence and performs local minimizations of
$\operatorname{KL}(p \gg q)$; this corresponds to iterative moment
matching on partitions of the data. Relative to \gls{klVI},
\gls{EP} produces overdispersed approximations. But \gls{EP} also has
drawbacks. It is not guaranteed to converge~\citep[Figure
$3.6$]{minka2001family}; it does not provide an easy estimate of the
marginal likelihood; and it does not optimize a well-defined global
objective~\citep{beal2003variational}.

In this paper we develop a new algorithm for approximate posterior
inference, \gls{chiVI}. \gls{chiVI} minimizes the $\chi$-divergence from the
posterior to the variational family,
\begin{equation}
  \vspace{-0.25ex}
  \label{eq:chi-divergence}
  D_{\chi^2}(p\gg q) =
  \mathbb{E}_{q(\mbz;
  \mblambda)}\Big[\Big(\frac{p(\mbz\g\mbx)}{q(\mbz; \mblambda)}\Big)^2
  - 1 \Big].
  \vspace{0.75ex}
\end{equation}
\gls{chiVI} enjoys advantages of both \gls{EP} and
\acrshort{klVI}. Like \gls{EP}, it produces overdispersed
approximations; like \gls{klVI}, it optimizes a well-defined objective
and estimates the model evidence. 

As we mentioned, \gls{klVI} optimizes a lower bound on the model
evidence. The idea behind \gls{chiVI} is to optimize an \emph{upper
  bound}, which we call the \gls{CUBO}. Minimizing the \gls{CUBO} is
equivalent to minimizing the $\chi$-divergence. 
In providing an upper bound, 
\gls{chiVI} can be used (in concert with \gls{klVI}) to
sandwich estimate the model evidence.
Sandwich estimates are
useful for tasks like model
selection~\citep{mackay1992bayesian}. Existing work on sandwich
estimation relies on MCMC and only evaluates simulated
data~\citep{grosse2015sandwiching}. 
We derive a \textit{sandwich theorem} (Section\nobreakspace \ref {sec:theory}) that relates \gls{CUBO} and \gls{ELBO}.
Section\nobreakspace \ref {sec:empirical} demonstrates sandwich estimation on real data.

Aside from providing an upper bound, 
there are two additional benefits to \gls{chiVI}. First, it
is a \textit{black-box inference algorithm}~\citep{ranganath2014black}
in that it does not need model-specific derivations 
and it is easy to apply to a wide class of models. 
It minimizes an upper bound in a principled way using unbiased reparameterization 
gradients~\citep{kingma2014autoencoding, rezende2014stochastic} of 
the exponentiated \gls{CUBO}. 

Second, it is a viable alternative to \gls{EP}.  The $\chi$-divergence
enjoys the same ``zero-avoiding'' behavior of \gls{EP}, which
seeks to place positive mass everywhere, and so \gls{chiVI}
is useful when the \gls{KL} divergence is not a good objective (such
as for light-tailed posteriors). 
Unlike \gls{EP}, \gls{chiVI} is guaranteed to converge; provides
an easy estimate of the marginal likelihood; and optimizes a
well-defined global objective.
Section\nobreakspace \ref {sec:empirical} shows that
\gls{chiVI} outperforms \gls{klVI} and \gls{EP} for Gaussian process classification.

The rest of this paper is organized as follows. Section\nobreakspace \ref {sec:theory}
derives the \gls{CUBO}, develops \gls{chiVI}, and expands on its
zero-avoiding property that finds overdispersed posterior
approximations. Section\nobreakspace \ref {sec:empirical} applies \gls{chiVI} to 
Bayesian probit regression, Gaussian process
classification, and a Cox process model of basketball plays. 
On Bayesian probit regression and Gaussian process 
classification, it yielded lower classification error than  
\gls{klVI} and \gls{EP}. When modeling basketball data with a Cox 
process, it gave more accurate estimates of posterior variance 
than \gls{klVI}.

\parhead{Related work.} The most widely studied variational objective is
$\operatorname{KL}(q\gg p)$.
The main alternative is \gls{EP} \cite{opper2000gaussian,minka2001family},
which locally minimizes
$\operatorname{KL}(p \gg q)$. Recent work revisits \gls{EP} from the
perspective of distributed
computing~\citep{gelman2014expectation,
teh2015distributed,
  li2015stochastic} and also revisits \cite{minka2004power}, which
studies local minimizations with the general family of
$\alpha$-divergences~\citep{hernandezlobato2015black,
  li2016variational}. \gls{chiVI} relates to \gls{EP} and its
extensions in that it leads to overdispersed approximations relative
to \gls{klVI}. However, unlike
\cite{minka2004power,hernandezlobato2015black}, \gls{chiVI} does not
rely on tying local factors; it optimizes a well-defined global
objective.
In this sense, \gls{chiVI} relates to the recent work on alternative
divergence measures for variational
inference~\citep{li2016variational, ranganath2016operator}.

A closely related work is \cite{li2016variational}. They perform
black-box variational inference using the reverse $\alpha$-divergence
$D_{\alpha}(q \gg p)$, which is a valid divergence
when $\alpha > 0$\footnote{It satisfies $D(p \gg q) \geq 0$ and $D(p \gg q) = 0 \iff p = q $ almost everywhere}.
Their work
shows that minimizing $D_{\alpha}(q \gg p)$ is equivalent to
maximizing a lower bound of the model evidence.  
No positive value of $\alpha$ in
$D_{\alpha}(q \gg p)$ leads to the $\chi$-divergence. 
Even though taking $\alpha \leq 0$
leads to \gls{CUBO}, 
it does not correspond to a valid divergence
in $D_{\alpha}(q \gg p)$.
The algorithm in \cite{li2016variational}
also cannot minimize the upper bound we study in this paper.
In this sense, our work complements \cite{li2016variational}.

An exciting concurrent work by \cite{kuleshov2017neural}
also studies the $\chi$-divergence. Their work focuses on upper bounding the partition
function in undirected graphical models. This is a complementary
application: Bayesian inference and undirected models both
involve an intractable normalizing constant.

\section{$\chi$-Divergence Variational Inference}
\label{sec:theory}
We present the $\chi$-divergence for variational inference. We describe
some of its properties and develop \gls{chiVI}, a black box algorithm that minimizes
the $\chi$-divergence for a large class of models.

\label{sec:background}
\Gls{VI} casts Bayesian inference as
optimization~\citep{jordan1999introduction}. 
\gls{VI} posits a family of approximating distributions and finds the closest
member to the posterior. In its typical formulation, \gls{VI} minimizes 
the Kullback-Leibler divergence from $q(\mbz; \mblambda)$ to $p(\mbz\g \mbx)$. 
Minimizing the KL divergence is equivalent to maximizing the 
\gls{ELBO}, a lower bound to the model evidence $\log p(\mbx)$. 

\vspace{-1.5ex}
\subsection{The $\chi$-divergence}
Maximizing the \gls{ELBO} imposes properties on the resulting
approximation such as underestimation of the posterior's
support~\citep{murphy2012machine, bishop2006pattern}.
These properties may be undesirable, especially when dealing
with light-tailed posteriors such as in Gaussian process 
classification~\citep{hensmantilted}. 

We consider
the $\chi$-divergence (Equation\nobreakspace \textup {\ref {eq:chi-divergence}}).
Minimizing the $\chi$-divergence induces
alternative properties on the resulting approximation.
(See Appendix\nobreakspace \ref {ap:properties} for more details on all these properties.)
Below we describe a key property
which leads to overestimation of the posterior's support.  
\vspace{-1.5ex}
\paragraph{Zero-avoiding behavior:}
Optimizing the $\chi$-divergence leads to 
a variational distribution with a \textit{zero-avoiding} behavior,
which is similar to EP \citep{minka2005divergence}.
Namely, the $\chi$-divergence is infinite whenever
$q(\mbz; \mblambda) = 0$ and $p(\mbz \g \mbx) > 0$.  
Thus when minimizing it, setting $p(\mbz \g \mbx) > 0$ forces
$q(\mbz; \mblambda) > 0$. This means $q$ 
avoids having zero mass at locations 
where $p$ has nonzero mass. 

The classical objective $\operatorname{KL}(q \gg p)$
leads to approximate posteriors with the opposite behavior, called \textit{zero-forcing}. 
Namely, $\operatorname{KL}(q \gg p)$ is infinite 
when $p(\mbz \g \mbx) = 0$ and
$q(\mbz; \mblambda) > 0$. Therefore the optimal variational
distribution $q$ will be $0$ when $p(\mbz \g \mbx) = 0$.
This \textit{zero-forcing} behavior leads
to degenerate solutions during optimization, and is the source of
``pruning'' often reported in the literature
(e.g., \citep{burda2016importance,hoffman2017learning}). For example, if the
approximating family $q$ has heavier tails than the target posterior
$p$, the variational distributions must be overconfident enough that
the heavier tail does not allocate mass outside the lighter tail's
support.\footnote{Zero-forcing may be preferable in settings such as
multimodal posteriors with unimodal approximations:
for predictive tasks, it helps to concentrate on one mode rather
than spread mass over all of them \citep{bishop2006pattern}. 
In this paper, we focus on
applications with light-tailed posteriors and one to relatively few
modes.}

\subsection{CUBO: the $\chi$ Upper Bound}
\label{sec:cubo}

We derive a tractable objective for variational inference
with the $\chi^2$-divergence
and also generalize it to the $\chi^n$-divergence for $n > 1$.
Consider the optimization problem of minimizing Equation\nobreakspace \textup {\ref {eq:chi-divergence}}. 
We seek to find a relationship between the $\chi^2$-divergence and $\log p(\mbx)$.
Consider
\begin{align*}
  \mathbb{E}_{q(\mbz; \mblambda)}\Big[\Big(\frac{p(\mbx, \mbz)}{q(\mbz; \mblambda)}\Big)^2 \Big]   
  	&= 1 + D_{\chi ^2}(p(\mbz|\mbx)\gg q(\mbz; \mblambda)) 
 	=  p(\mbx)^2[1 + D_{\chi ^2}(p(\mbz| \mbx)\gg q(\mbz; \mblambda))].
\end{align*}
Taking logarithms on both sides, we find a relationship
analogous to how 
$\operatorname{KL}(q \gg p)$ relates to the \gls{ELBO}.  Namely, the
$\chi^2$-divergence satisfies
\begin{align*}
 \frac{1}{2} &\log(1 + D_{\chi ^2}(p(\mbz|\mbx)\gg q(\mbz; \mblambda))) =
 -\log p(\mbx) + \frac{1}{2}\log\mathbb{E}_{q(\mbz; \mblambda)}
 	\Big[\Big(\frac{p(\mbx,\mbz)}{q(\mbz; \mblambda)}\Big)^2 \Big].
\label{eq:cubo_identity}
\end{align*}
By monotonicity of $\log$, and because $\log p(\mbx)$ is constant,
 minimizing the $\chi^2$-divergence is equivalent to minimizing
\begin{equation*}
\mathcal{L}_{\chi^2}(\mblambda) = \frac{1}{2}\log\mathbb{E}_{q(\mbz; \mblambda)}
			 		\Big[\Big(\frac{p(\mbx,\mbz)}{q(\mbz; \mblambda)}\Big)^2 \Big]
.
\end{equation*}
Furthermore, by nonnegativity of the $\chi^2$-divergence, this quantity 
is an upper bound to the model evidence. 
We call this objective the \glsreset{CUBO}\emph{\gls{CUBO}}.

\parhead{A general upper bound.}
The derivation extends to upper bound the general
$\chi^n$-divergence,
\begin{equation}
 \label{eq:family_bounds}
 \mathcal{L}_{\chi^n}(\mblambda) = 
 	\frac{1}{n}\log\mathbb{E}_{q(\mbz; \mblambda)}
        \Big[\Big(\frac{p(\mbx,\mbz)}{q(\mbz; \mblambda)}\Big)^n \Big] = \gls{CUBO}_n .
\end{equation}
This produces a family of bounds.
When $n < 1$, \gls{CUBO}$_n$ is a lower bound, and 
minimizing it for these values of $n$ does not minimize 
the $\chi$-divergence (rather, when $n < 1$, we recover the reverse 
$\alpha$-divergence and the VR-bound~\citep{li2016variational}).
When $n=1$, the bound is tight where \gls{CUBO}$_1 =\log p(\mbx)$. 
For $n \geq 1$, \gls{CUBO}$_n$ is an upper bound to the model evidence. 
In this paper we focus on $n=2$. Other values of $n$ are possible depending on 
the application and dataset. We chose $n=2$ because it is the most
standard, and is equivalent to finding the optimal proposal in importance sampling. 
See Appendix\nobreakspace \ref {ap:is} for more details. 

\parhead{Sandwiching the model evidence.} Equation\nobreakspace \textup {\ref {eq:family_bounds}} has
practical value. We can minimize the
\gls{CUBO}$_n$ and maximize the \gls{ELBO}. This produces a sandwich
on the model evidence. 
(See Appendix\nobreakspace \ref {ap:simulations} for a simulated illustration.)
The following \textit{sandwich theorem} states that the gap
induced by \gls{CUBO}$_n$ and \gls{ELBO} increases
with $n$. This suggests that letting $n$ as close to $1$ as possible
enables approximating $\log p(\mbx)$
with higher precision. When we further 
decrease $n$ to $0$,  \gls{CUBO}$_n$ becomes 
a lower bound and tends to the \gls{ELBO}.

\begin{theorem} \label{eq:theorem}
\emph{\textbf{(Sandwich Theorem)}:}
Define $\gls{CUBO}_n$ as in~Equation\nobreakspace \textup {\ref {eq:family_bounds}}.  Then the following holds: 
\begin{itemize}
	\item $\forall n \geq 1$ $\gls{ELBO} \leq \log p(\mbx) \leq \gls{CUBO}_n$. 
	\item $\forall n \geq 1$ $\gls{CUBO}_n$ is a non-decreasing function of the order $n$ of the $\chi$-divergence.
	\item $\lim_{n \to 0} \gls{CUBO}_n = \gls{ELBO} $.
\end{itemize}
\end{theorem}

See proof in Appendix\nobreakspace \ref {ap:proof}.
Theorem\nobreakspace \ref {eq:theorem} can be utilized for estimating $\log p(\mbx)$, which is important for
many applications such as the evidence
framework~\citep{mackay2003information}, where the marginal
likelihood is argued to embody an Occam's razor. 
Model selection based solely on the \gls{ELBO} is inappropriate 
because of the possible variation in the tightness of this bound. 
With an accompanying upper bound, one can perform what we call 
\textit{maximum entropy model selection} in which each model 
evidence values are chosen to be that which maximizes the 
entropy of the resulting distribution on models. We leave this as future work.
Theorem\nobreakspace \ref {eq:theorem} can also help estimate Bayes factors~\citep{raftery1995bayesian}.
In general, this technique is important as there is little existing work: for example, Ref. \cite{grosse2015sandwiching} proposes an
\acrshort{MCMC} approach and evaluates simulated
data. 
We illustrate sandwich estimation in Section\nobreakspace \ref {sec:empirical} on UCI datasets.
 \subsection{Optimizing the \gls{CUBO}}
\label{sec:optimizing}
We derived the \gls{CUBO}$_n$, a general upper bound to the model evidence
that can be used to minimize the $\chi$-divergence. We now develop
\gls{chiVI}, a black box algorithm that minimizes \gls{CUBO}$_n$.

The goal in \gls{chiVI} is to minimize the \gls{CUBO}$_n$ with respect
to variational parameters,
\begin{equation*}
\gls{CUBO}_n(\mblambda) = \frac{1}{n}\log\mathbb{E}_{q(\mbz; \mblambda)}
	\Big[\Big(\frac{p(\mbx,\mbz)}{q(\mbz ;  \mblambda)}\Big)^n \Big].
\end{equation*}
The expectation in the \gls{CUBO}$_n$ is usually intractable. 
Thus we use Monte Carlo to construct an estimate. One approach is to
naively perform Monte Carlo on this objective,
\begin{equation*}
\gls{CUBO}_n(\mblambda) \approx \frac{1}{n}\log \frac{1}{S}\sum_{s=1}^S 
	\Big[\Big(\frac{p(\mbx,\mbz^{(s)})}{q(\mbz^{(s)} ;  \mblambda)}\Big)^n \Big],
\end{equation*}
for $S$ samples $\mbz^{(1)}, ...,\mbz^{(S)}\sim q(\mbz; \mblambda)$.
However, by Jensen's inequality, the $\log$ transform of the expectation implies that
this is a biased estimate of $\gls{CUBO}_n(\mblambda)$:
\begin{equation*}
\mathbb{E}_q\Bigg[\frac{1}{n}\log \frac{1}{S}\sum_{s=1}^S 
	\Big[\Big(\frac{p(\mbx,\mbz^{(s)})}{q(\mbz^{(s)} ;  \mblambda)}\Big)^n \Big]\Bigg] \ne \gls{CUBO}_n. 
\end{equation*}
In fact this expectation changes during optimization and depends 
on the sample size $S$. The objective is not guaranteed to be an upper 
bound if $S$ is not chosen appropriately from the beginning. 
 This problem does not exist for lower bounds because the Monte Carlo approximation 
is still a lower bound; this is why the approach 
in~\citep{li2016variational} works for lower bounds 
but not for upper bounds. Furthermore, gradients of this 
biased Monte Carlo objective are also biased. 

We propose a way to minimize upper bounds which also can be used for lower bounds. 
The approach keeps the upper bounding property intact. 
It does so by minimizing a Monte Carlo approximation of the 
exponentiated upper bound,
\begin{equation*}
\mbL = \exp\{n\cdot \gls{CUBO}_n(\mblambda)\}
.
\end{equation*} 
By monotonicity of $\exp$,
this objective admits the same optima as $\gls{CUBO}_n(\mblambda)$.
Monte Carlo produces an unbiased estimate,
and the number of samples 
only affects the variance of the gradients.
We minimize it using reparameterization
gradients~\citep{kingma2014autoencoding,rezende2014stochastic}.
These gradients apply to models with differentiable latent variables.
Formally, assume we can rewrite the generative process as $\mbz =
g(\mblambda, \epsilon)$ where $\epsilon \sim p(\epsilon)$ and for some
deterministic function $g$.
Then
\begin{equation*}
 \hat{\mbL} = \frac{1}{B} \sum_{b=1}^{B}  \Big(\frac{p(\mbx, g(\mblambda, \epsilon^{(b)}))}
 {q(g(\mblambda, \epsilon^{(b)}) ; \mblambda)}\Big)^n
\end{equation*}
is an unbiased estimator of $\mbL$ and its gradient is
\begin{align}
\label{eq:grad}
 \nabla_{\mblambda}  \hat{\mbL}
 &= \frac{n}{B} \sum_{b=1}^{B} \Big(\frac{p(\mbx, g(\mblambda, \epsilon^{(b)}))}
 	{q(g(\mblambda, \epsilon^{(b)}) ; \mblambda)}\Big)^n
\nabla_{\mblambda}\log\Big(\frac{p(\mbx, g(\mblambda, \epsilon^{(b)}))}
{q(g(\mblambda, \epsilon^{(b)}) ; \mblambda)} \Big)
.
\end{align}
(See Appendix\nobreakspace \ref {ap:fdiv} for a more detailed derivation 
and also a more general alternative with
score function gradients~\citep{paisley2012variational}.)

Computing Equation\nobreakspace \textup {\ref {eq:grad}} requires the full dataset $\mbx$.
We can apply the ``average likelihood'' technique from
\gls{EP}~\citep{li2015stochastic,dehaene2015expectation}.
Consider data $\{\mbx_1,\ldots, \mbx_N\}$ and a subsample $\{\mbx_{i_1}, ... , \mbx_{i_M}\}$..
We approximate the full log-likelihood by 
\begin{equation*}
\log p(\mbx \g \mbz) \approx  \frac{N}{M}\sum_{j=1}^{M} \log p(\mbx_{i_j} \g  \mbz).
\end{equation*}
Using this proxy to the full dataset we derive \gls{chiVI},
 an algorithm in which each iteration 
depends on only a mini-batch of data. 
\gls{chiVI} is a black box algorithm for performing
approximate inference with the $\chi^n$-divergence. 
Algorithm\nobreakspace \ref {alg:scalable_chivi} summarizes the procedure.
In practice, we subtract the maximum of the logarithm 
of the importance weights, defined as
\begin{equation*}
	\log\mbw = \log p(\mbx,\mbz) - \log q(\mbz; \mblambda).
\end{equation*}
to avoid underflow. 
Stochastic optimization theory still gives us convergence 
with this approach~\citep{sunehag2009variable}.
\begin{algorithm}[t]
 \caption{\glsreset{chiVI}\gls{chiVI}}
  \SetAlgoLined
  \DontPrintSemicolon
  \BlankLine
  \KwIn{Data $\mbx$, Model $p(\mbx,\mbz)$, Variational family $q(\mbz;
  \mblambda)$.}
  \BlankLine
  \textbf{Output}: Variational parameters $\mblambda$.\;
  \BlankLine
  Initialize $\mblambda$ randomly.
  \BlankLine
  \While{\textnormal{not converged}}{
    \BlankLine
    Draw $S$ samples $\mbz^{(1)},...,\mbz^{(S)}$ from $q(\mbz; \mblambda)$ 
    and a data subsample $\{x_{i_1}, ... , x_{i_M}\}$.
    \BlankLine
    Set $\rho_t$ according to a learning rate schedule.
    \BlankLine
     Set $\log \mbw^{(s)} = \log p(\mbz^{(s)}) + \frac{N}{M} \sum_{j=1}^{M} p(\mbx_{i_j} \g  \mbz)
     - \log q(\mbz^{(s)}; \mblambda_t)$, $s\in \{1, ..., S\}$.
    \BlankLine
    Set $\mbw^{(s)} =\exp( \log \mbw^{(s)}  - \displaystyle\max_{s} \log \mbw^{(s)}), s\in \{1, ..., S\}$.
    \BlankLine
    \vspace{-1.0ex}
    Update
$\mblambda_{t+1} = \mblambda_{t} - \frac{(1-n)\cdot \rho_t}{S} \sum_{s = 1}^{S}
		\Big[\Big(\mbw^{(s)}\Big)^n
		\nabla_{\mblambda}\log q(\mbz^{(s)}; \mblambda_t)\Big]$.
  }
  \label{alg:scalable_chivi}
    \vspace{-1.0ex}
\end{algorithm}

 \vspace{-2ex}
\section{Empirical Study}
\label{sec:empirical}
We developed \gls{chiVI}, a black box variational inference 
algorithm for minimizing the $\chi$-divergence. 
We now study \gls{chiVI} with several models: probit regression, 
\gls{GP} classification, and Cox processes. 
With probit regression, we demonstrate the sandwich estimator on 
real and synthetic data. \gls{chiVI} provides a useful tool to
estimate the marginal likelihood.
We also show that for this model where \gls{ELBO} is 
applicable \gls{chiVI} works well and yields good test error rates. 

Second, we compare \gls{chiVI} to Laplace and \gls{EP} on 
GP classification, a model class for which \gls{klVI} fails (because 
the typical chosen variational distribution has heavier tails 
than the posterior).\footnote{For \gls{klVI}, we use the \gls{BBVI} version \cite{ranganath2014black}
specifically via Edward \citep{tran2016edward}.}
In these settings, \gls{EP} has been the 
method of choice. \gls{chiVI} outperforms both of these methods.

Third, we show that \gls{chiVI} does not suffer from the posterior support 
 underestimation problem resulting from maximizing the \gls{ELBO}. 
For that we analyze Cox processes, a type of spatial point process, to
compare profiles of different NBA basketball players.
We find \gls{chiVI} yields better posterior uncertainty 
estimates (using HMC as the ground truth). 

\subsection{Bayesian Probit Regression}
\begin{figure*}[t]
\centering
\includegraphics[scale=0.21]{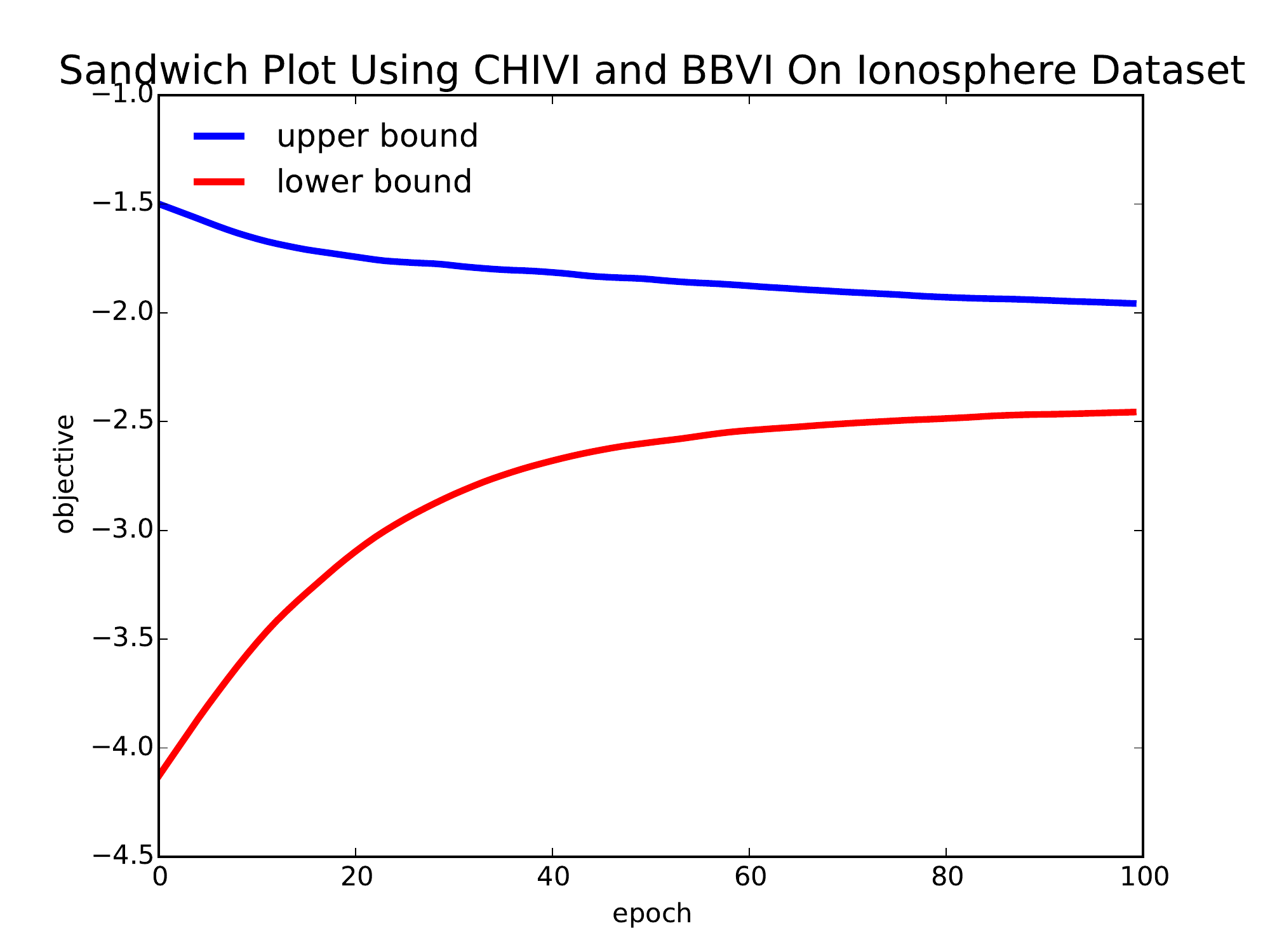}
\includegraphics[scale=0.21]{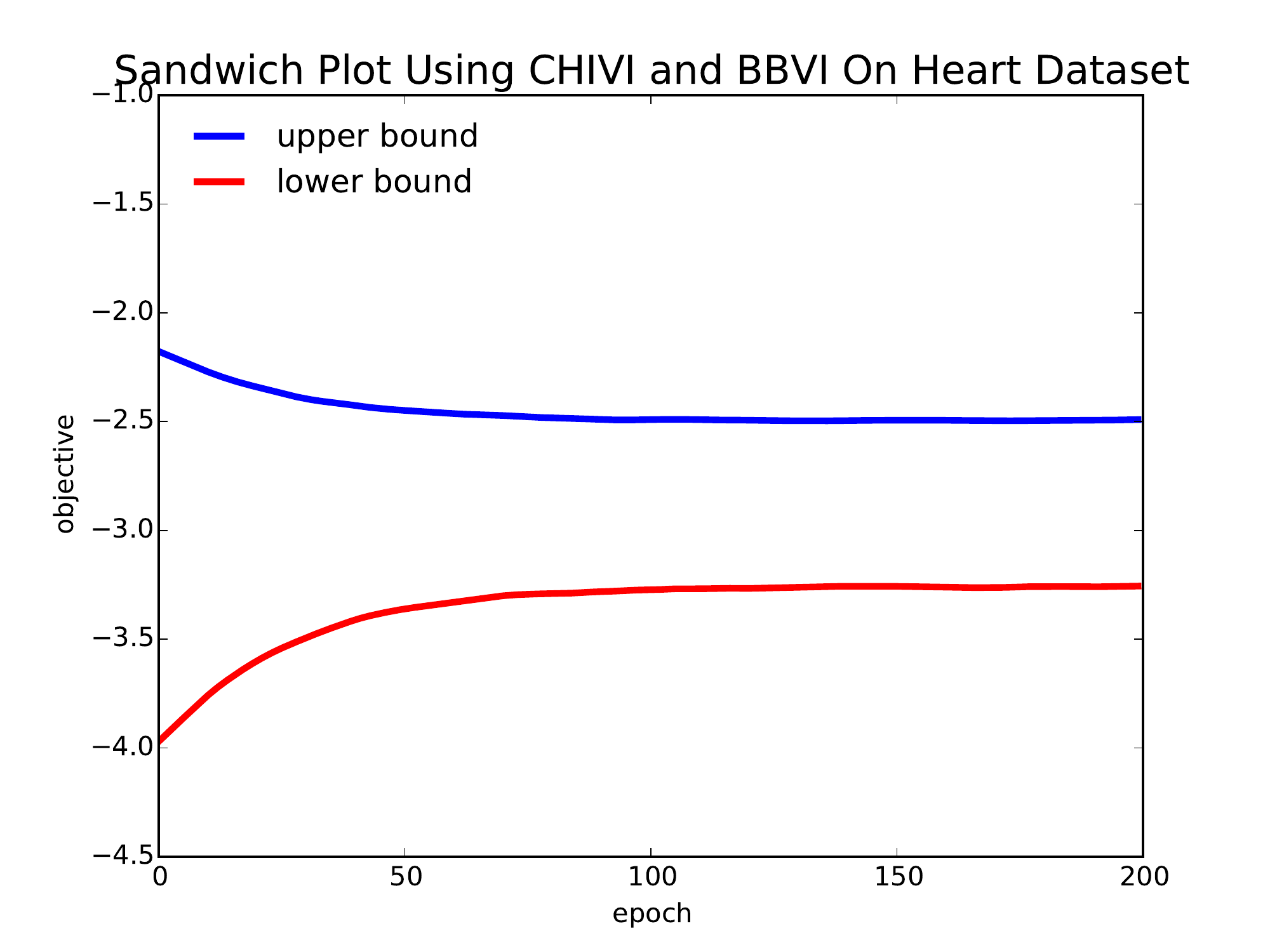}
\includegraphics[scale=0.252]{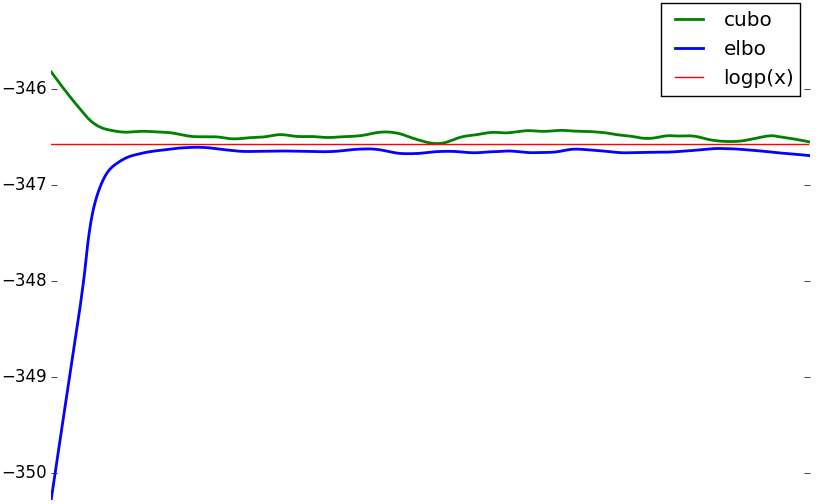}
\caption{Sandwich gap via \gls{chiVI} and \gls{BBVI} on different datasets.
The first two plots correspond to 
sandwich plots for the two UCI datasets \emph{Ionosphere} 
and \emph{Heart} respectively.  
 The last plot corresponds to a sandwich for generated 
 data where we know the log marginal likelihood of 
 the data. There the gap is tight after only few iterations. 
 More sandwich plots can be found in the appendix.}
\label{fig:sandwich}
\end{figure*}

We analyze inference for Bayesian probit regression.
First, we illustrate sandwich estimation on UCI datasets.  
Figure\nobreakspace \ref {fig:sandwich} illustrates the bounds of the
log marginal likelihood given by the \gls{ELBO} and the \gls{CUBO}.
Using both quantities provides a reliable approximation of the
model evidence. 
In addition, these figures show convergence 
for \gls{chiVI}, which \gls{EP} does not always satisfy.

We also compared the predictive performance 
of \gls{chiVI}, \gls{EP}, and \gls{klVI}.  
We used a minibatch size of $64$ and $2000$ iterations for each batch. 
We computed the average classification error
rate and the standard deviation using $50$ random splits of the data.
We split all the datasets with $90\%$ of the
data for training and $10\%$ for testing. 
For the Covertype dataset, we implemented
Bayesian probit regression to discriminate the class $1$ against all
other classes. Table\nobreakspace \ref {tab:error_probit} shows the average error rate
for \gls{klVI},
\gls{EP}, and \gls{chiVI}. \gls{chiVI} performs better
for all but one dataset.

\begin{table}[!t]
\vspace{-1ex}
\caption{Test error for Bayesian probit regression. The lower the better. 
\gls{chiVI} (this paper) yields lower test error rates 
when compared to \gls{BBVI}~\citep{ranganath2014black}, and \gls{EP} on most datasets.\\}
\centering
\begin{tabular}[\textwidth]{ccccccc}
\toprule
Dataset & \gls{BBVI} & \gls{EP} & \gls{chiVI} \\
\midrule
Pima &  0.235 $\pm$ 0.006 &  0.234 $\pm$ 0.006 & \textbf{0.222 $\pm$ 0.048}  \\
Ionos & 0.123 $\pm$ 0.008 & 0.124 $\pm$ 0.008 &\hspace{-0.75em} \textbf{0.116 $\pm$ 0.05} \\
Madelon & 0.457 $\pm$ 0.005 & \textbf{0.445 $\pm$ 0.005} & 0.453 $\pm$ 0.029 \\
Covertype &  0.157 $\pm$ 0.01  &  0.155 $\pm$ 0.018 & \textbf{0.154 $\pm$ 0.014} \\\bottomrule
\end{tabular}
\label{tab:error_probit}
\end{table}

\subsection{Gaussian Process Classification}
\begin{table}[!hbpt]
\caption{Test error for \acrlong{GP} classification. The lower the better.
\gls{chiVI} (this paper) yields lower test error rates 
when compared to Laplace and \gls{EP} on most datasets.\\}
\centering
\begin{tabular}{cccc}
\toprule
Dataset & Laplace & \gls{EP} & \gls{chiVI} \\
\midrule
Crabs & \textbf{0.02} & \textbf{0.02} & 0.03 $\pm$ 0.03\\
Sonar & 0.154 & 0.139 & \textbf{0.055 $\pm$ 0.035} \\
Ionos & 0.084 & 0.08 $\pm$ 0.04 & \textbf{0.069 $\pm$ 0.034}\\\bottomrule
\end{tabular}
\label{table:error_rates}
\vspace{-2ex}
\end{table}

\begin{figure*}[t]
   \centering
    {\includegraphics[scale=0.21]{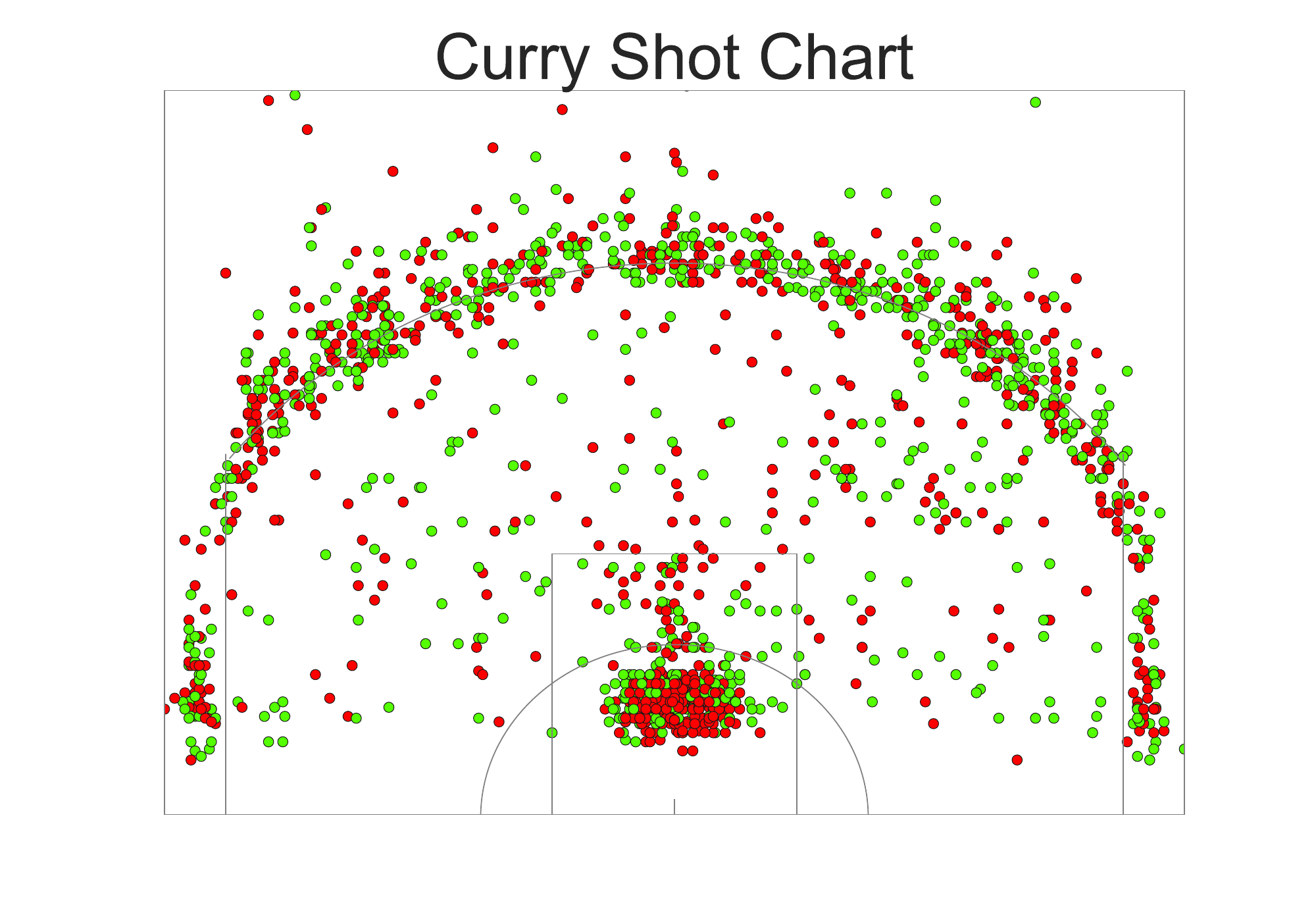}}
    {\includegraphics[scale=0.21]{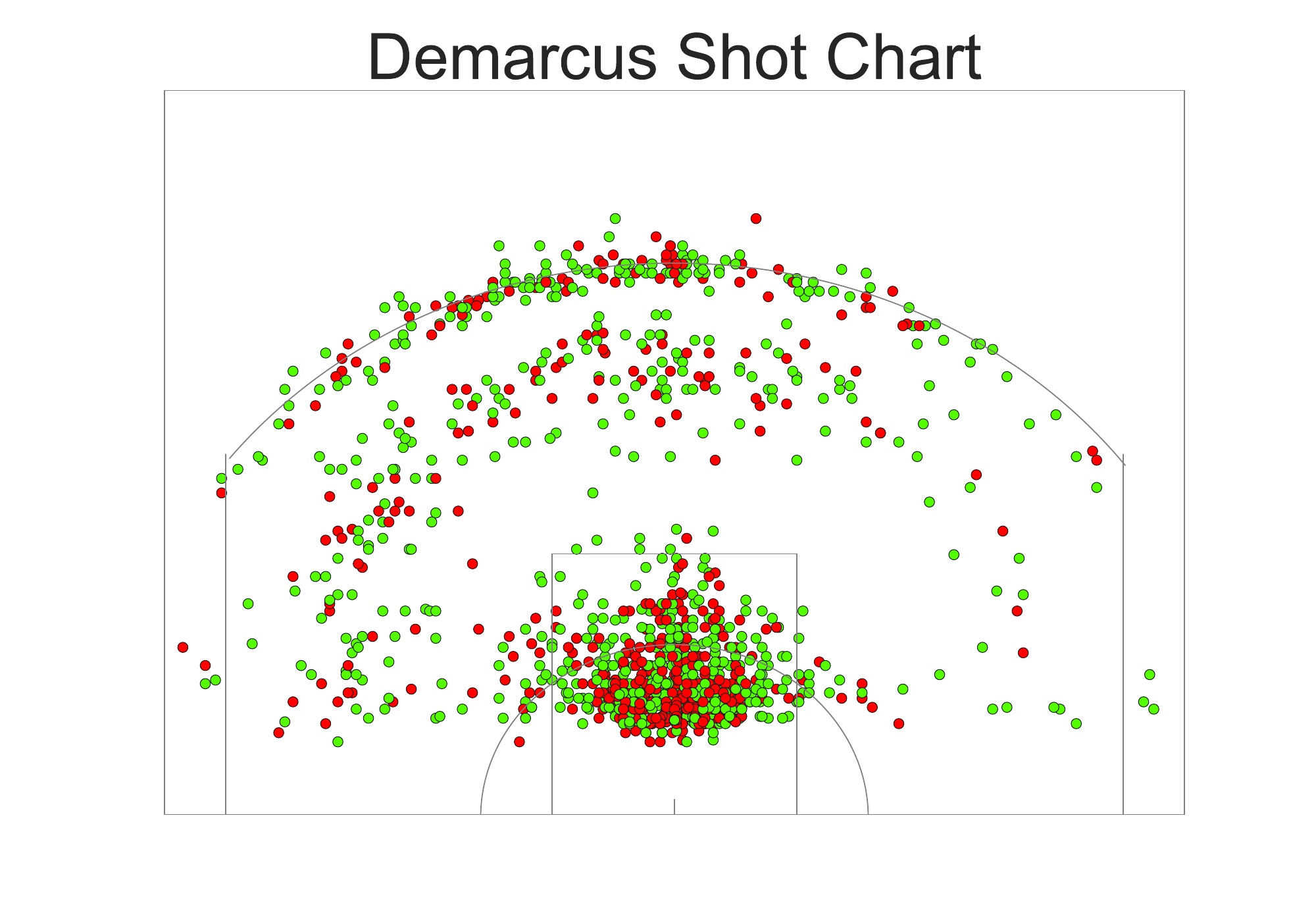}} \\
    {\includegraphics[scale=0.21]{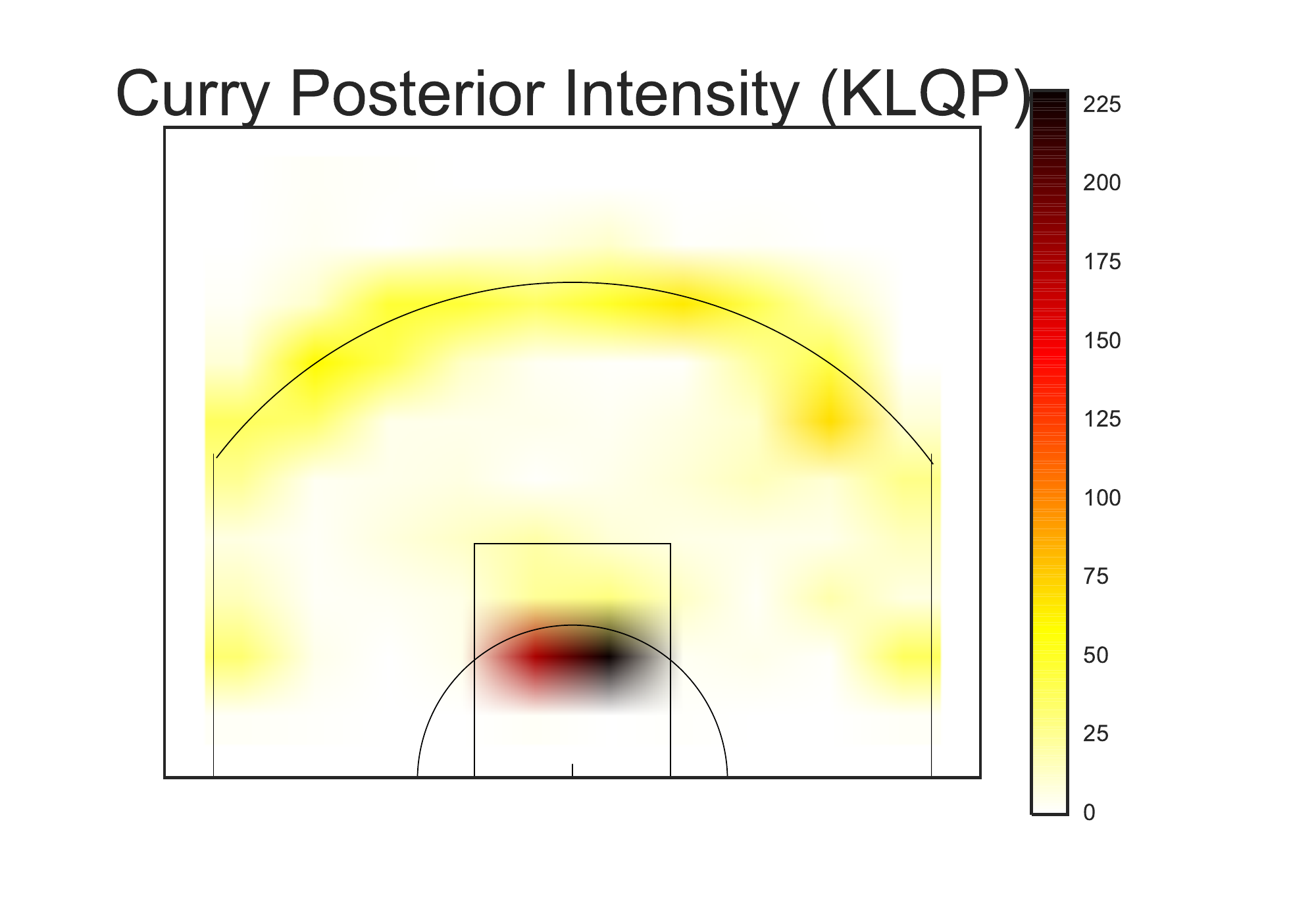}}
    {\includegraphics[scale=0.21]{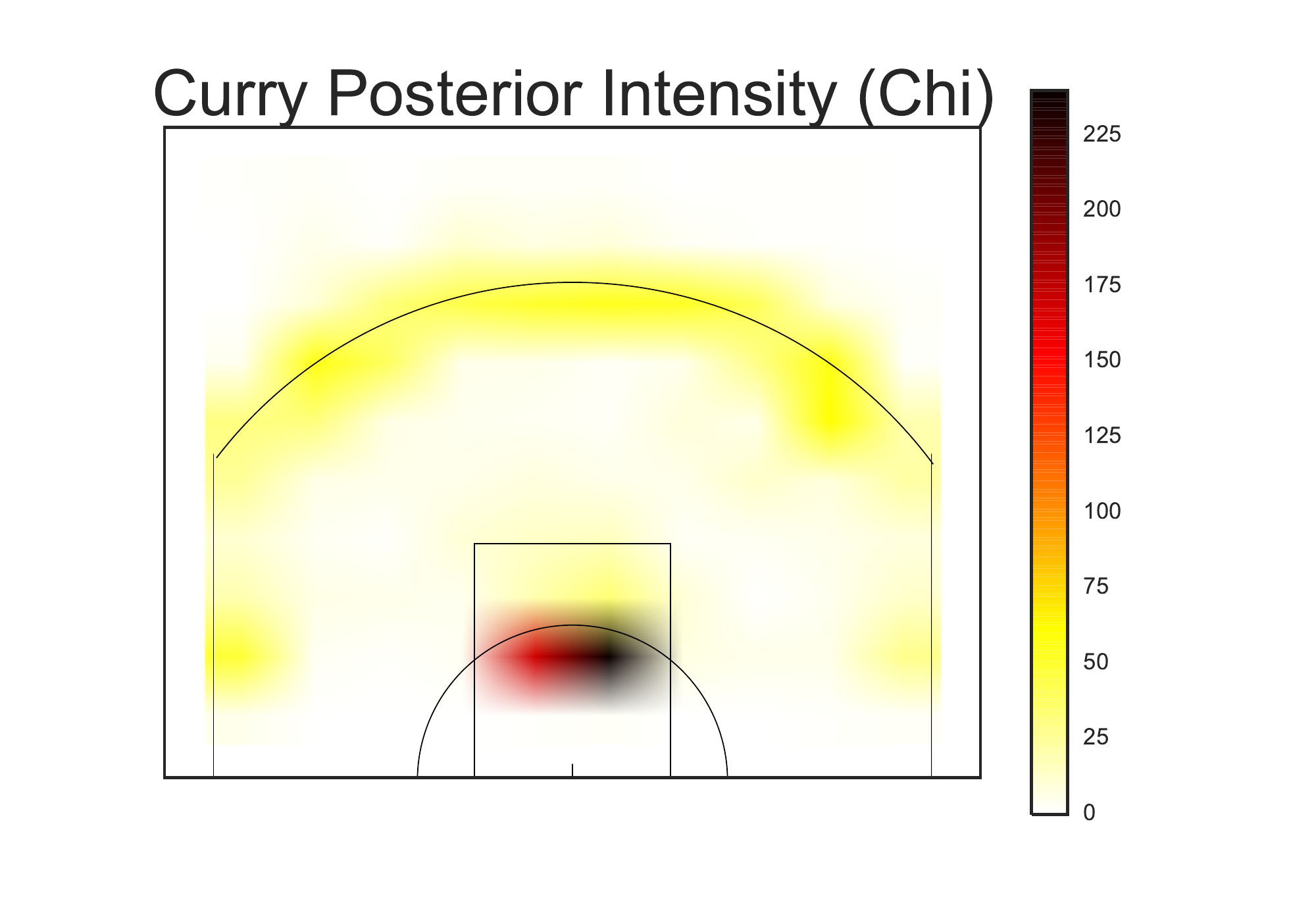}}
    {\includegraphics[scale=0.21]{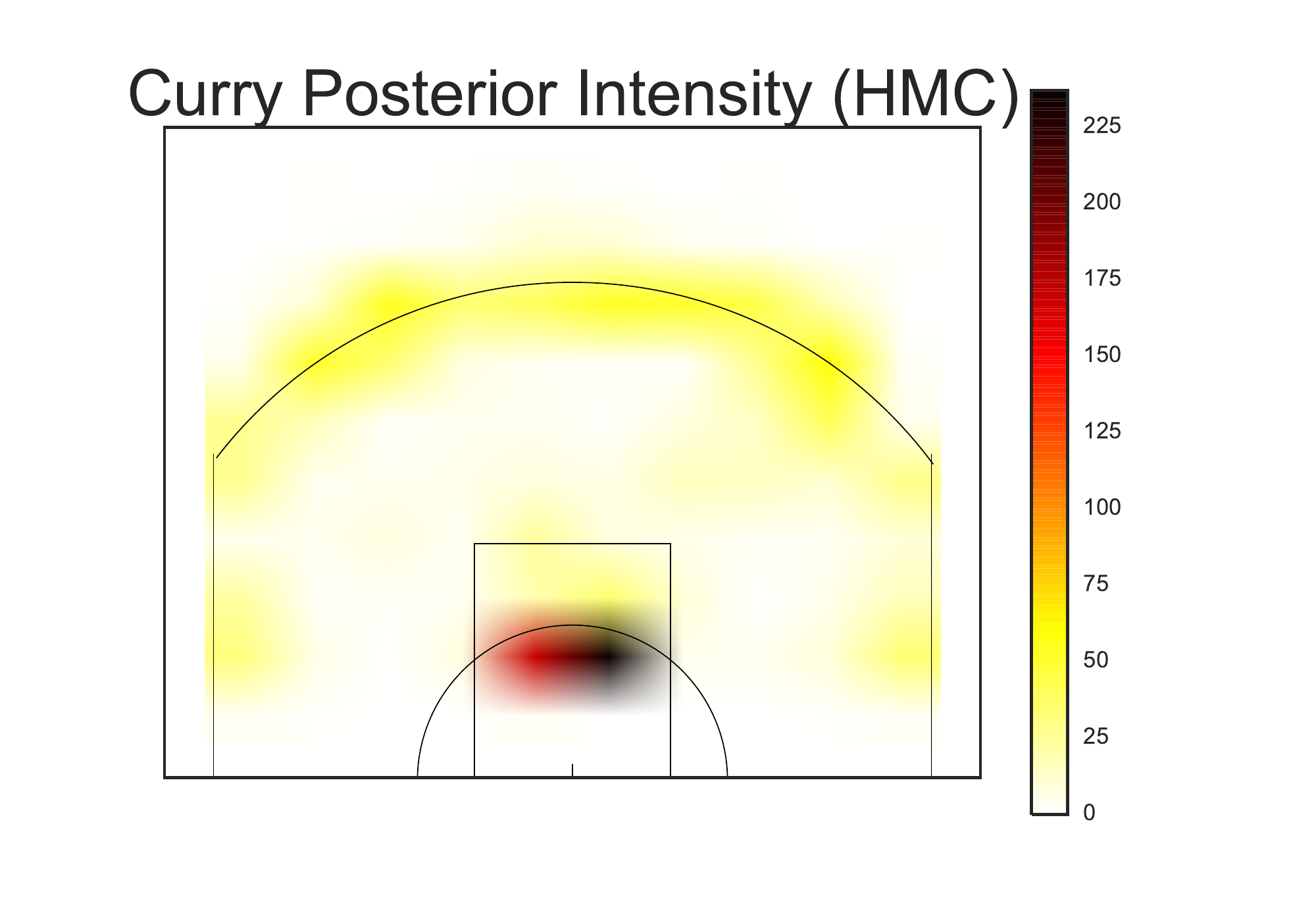}}\\
    {\includegraphics[scale=0.21]{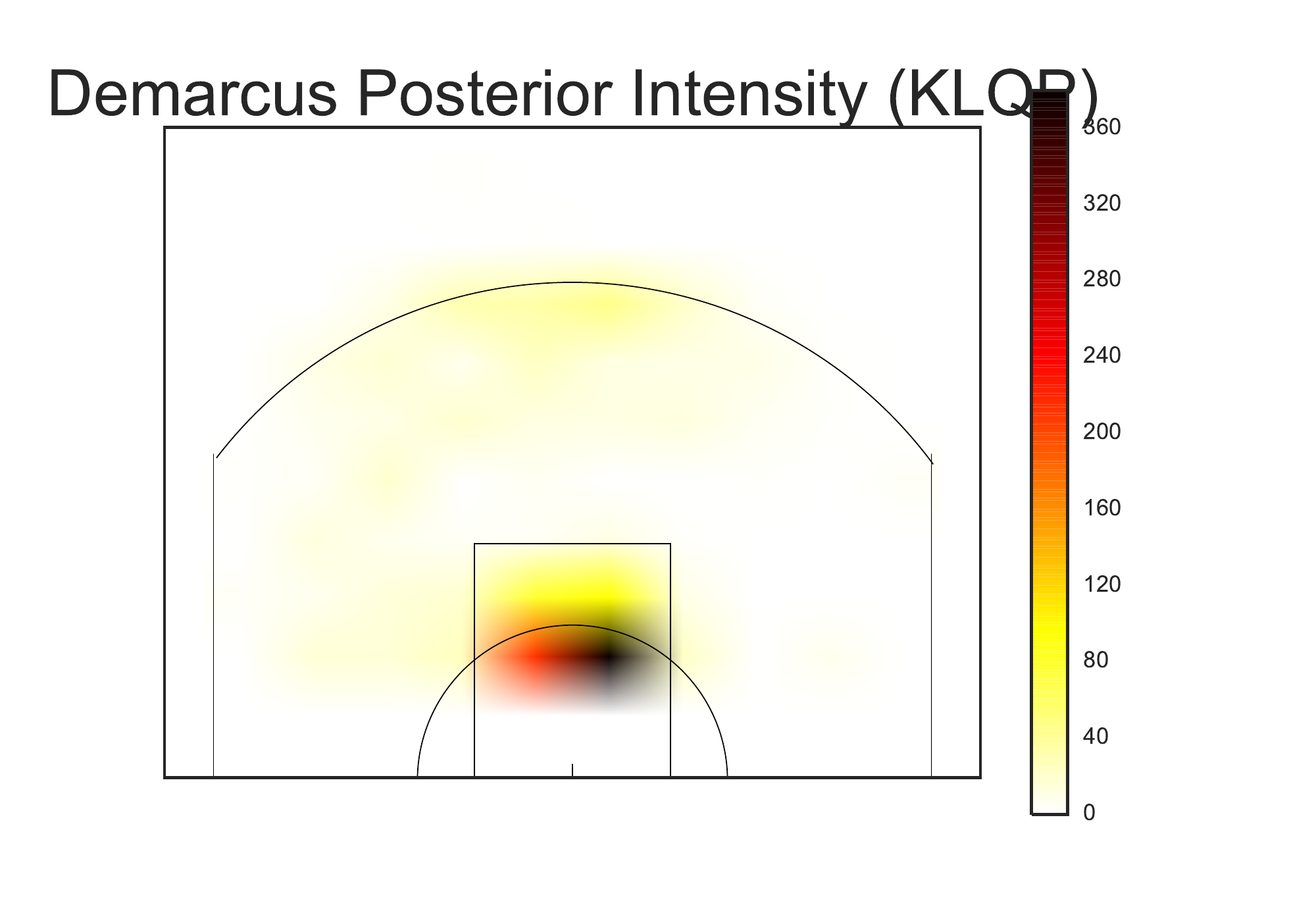}}
    {\includegraphics[scale=0.21]{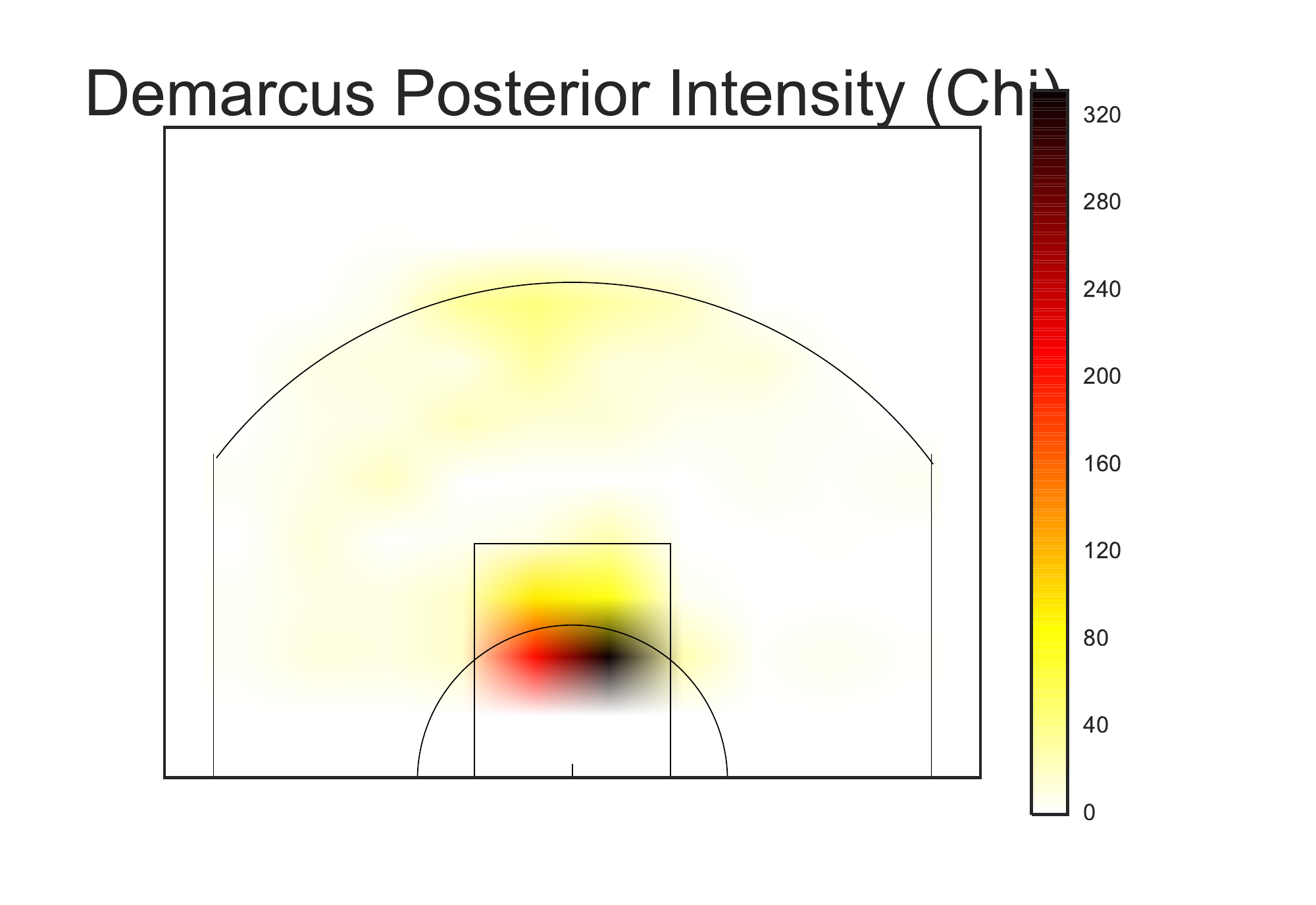}}
    {\includegraphics[scale=0.21]{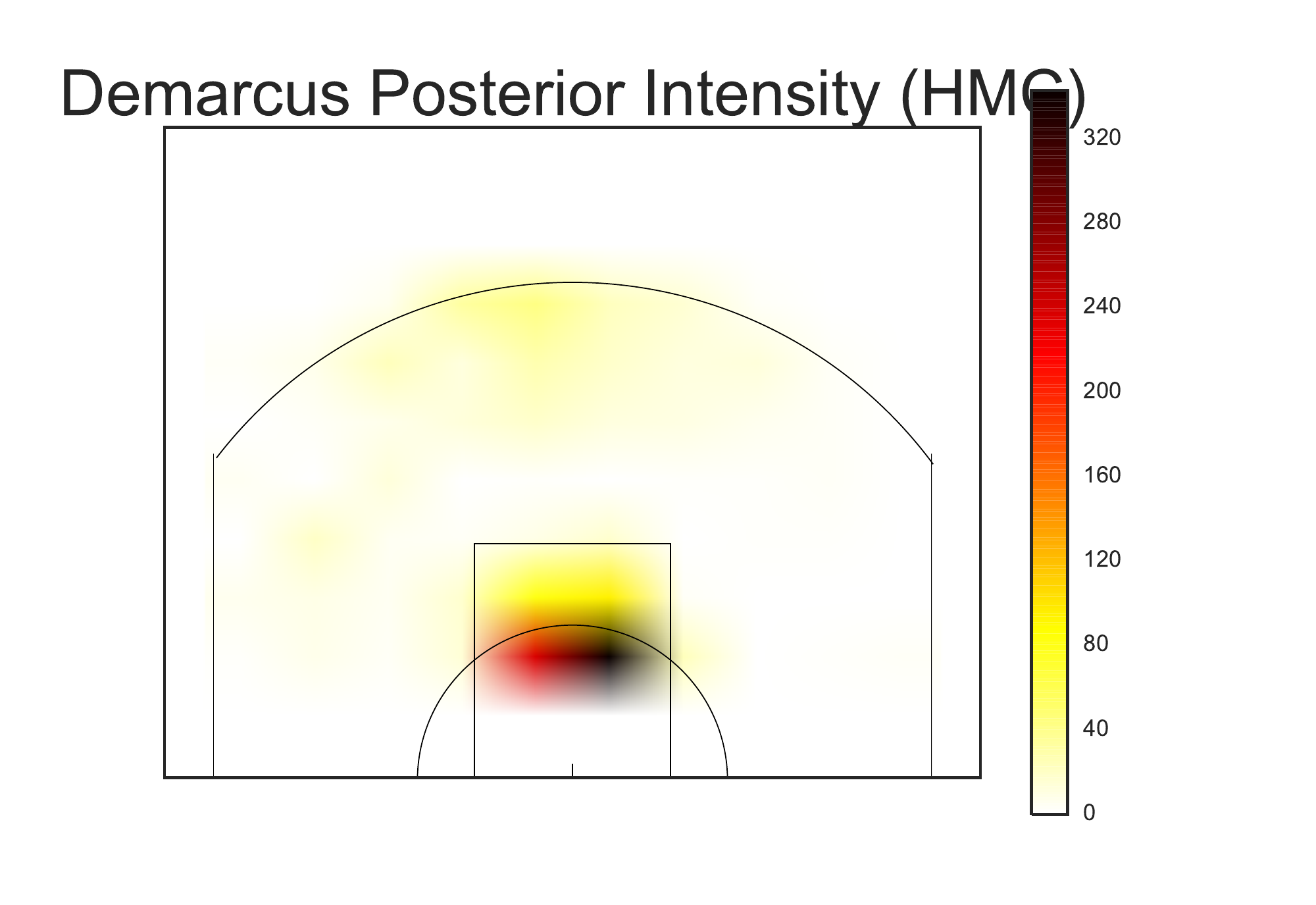}}\\
     {\includegraphics[scale=0.21]{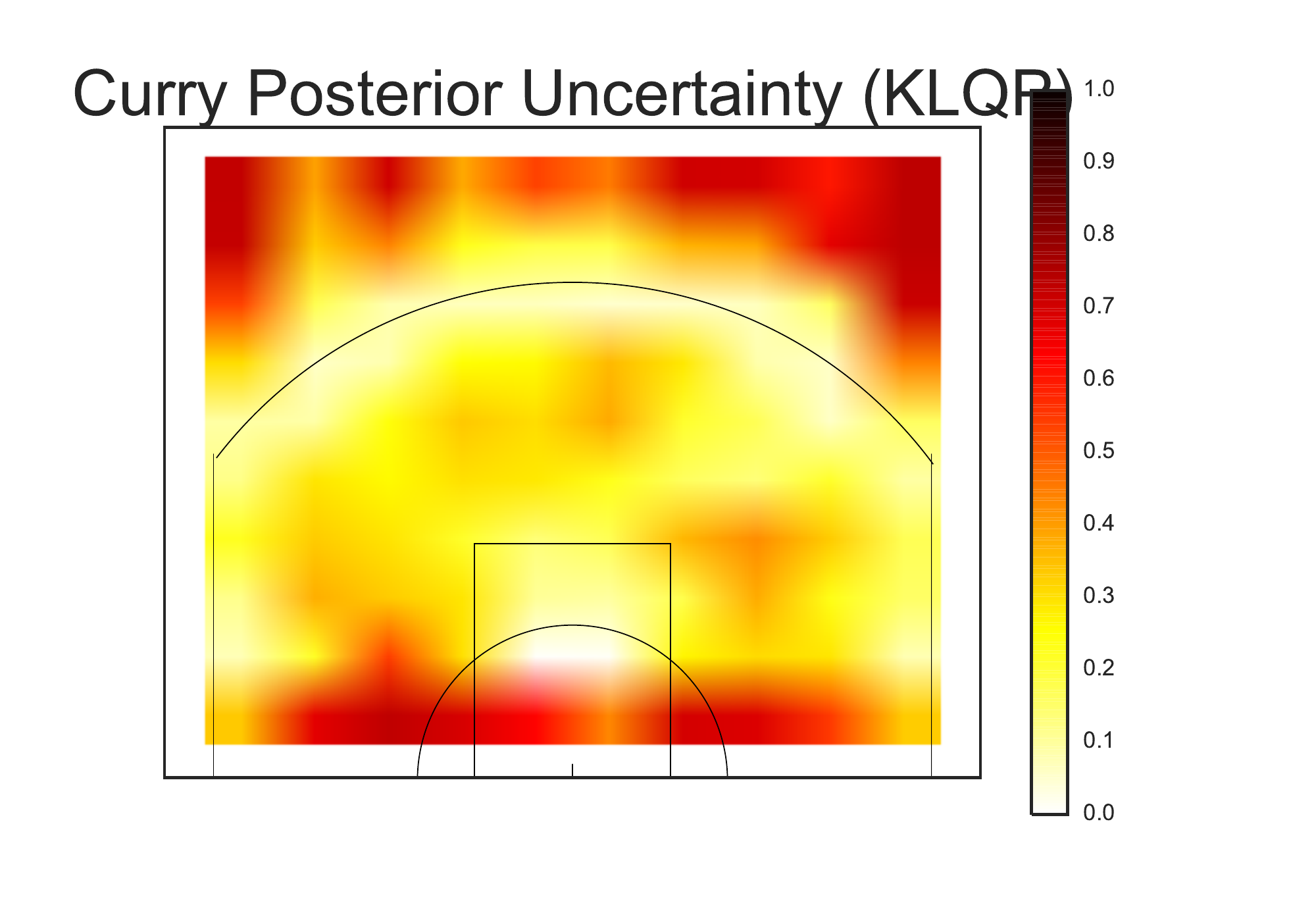}}
     {\includegraphics[scale=0.21]{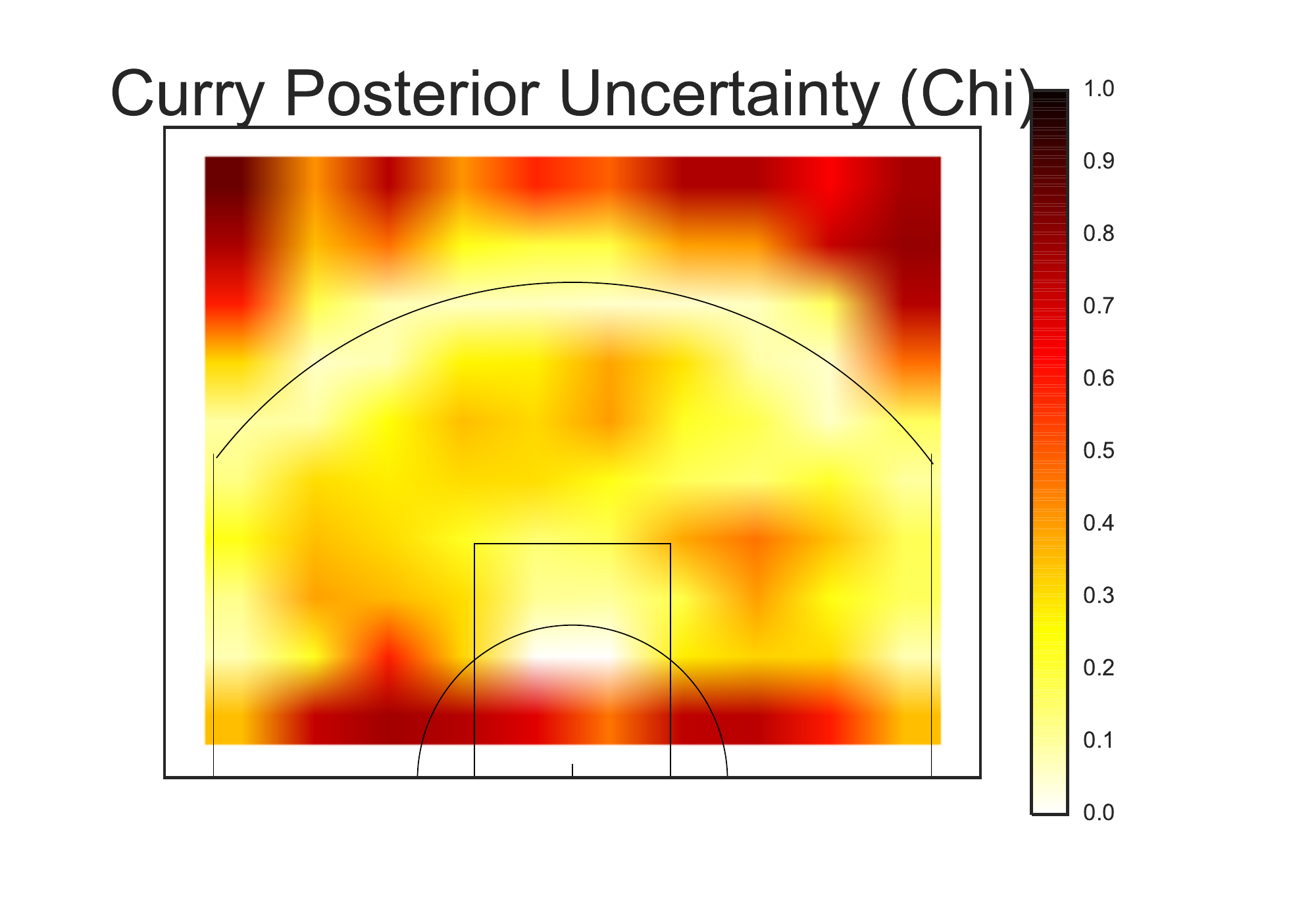}}
     {\includegraphics[scale=0.21]{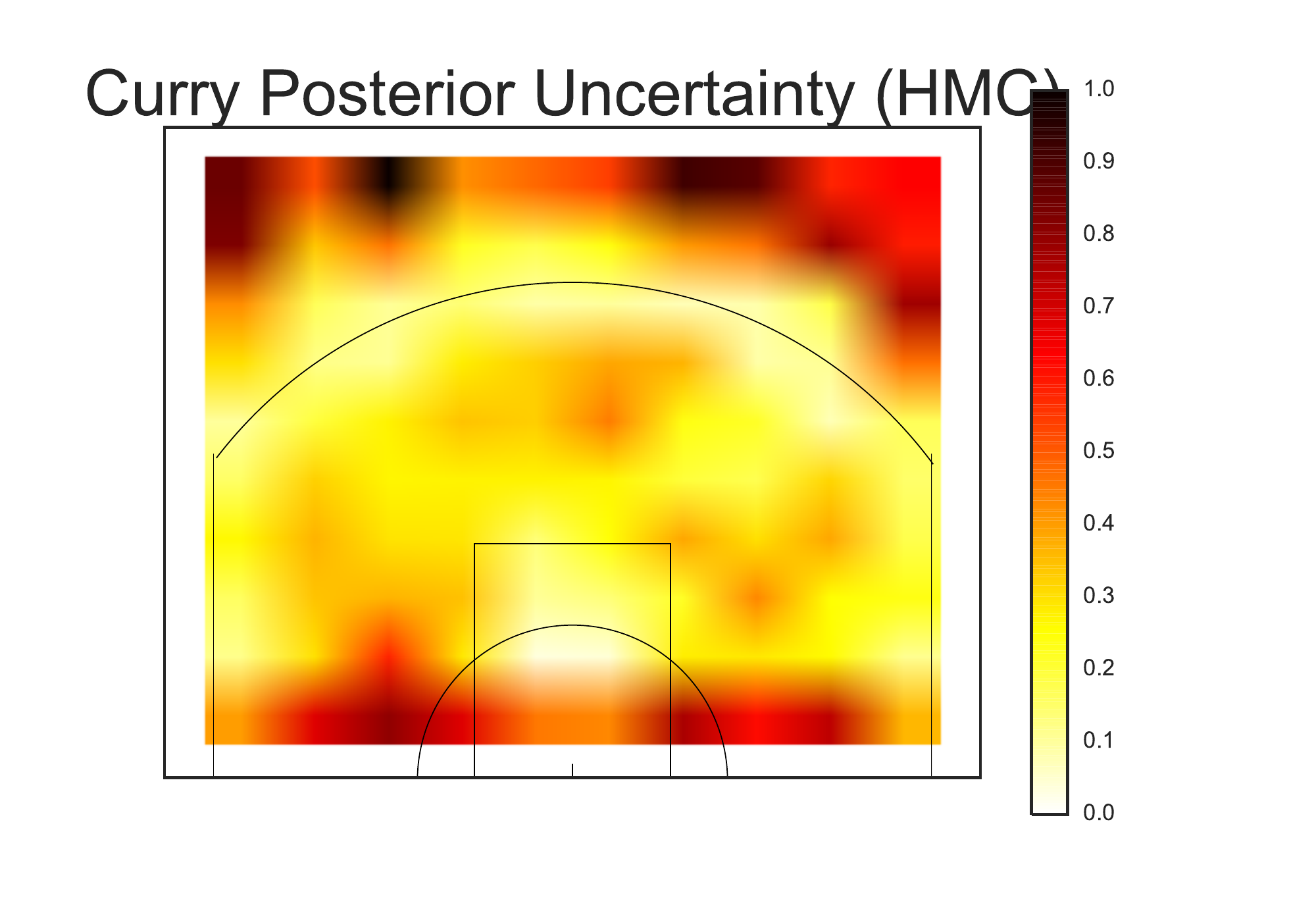}}\\
    {\includegraphics[scale=0.21]{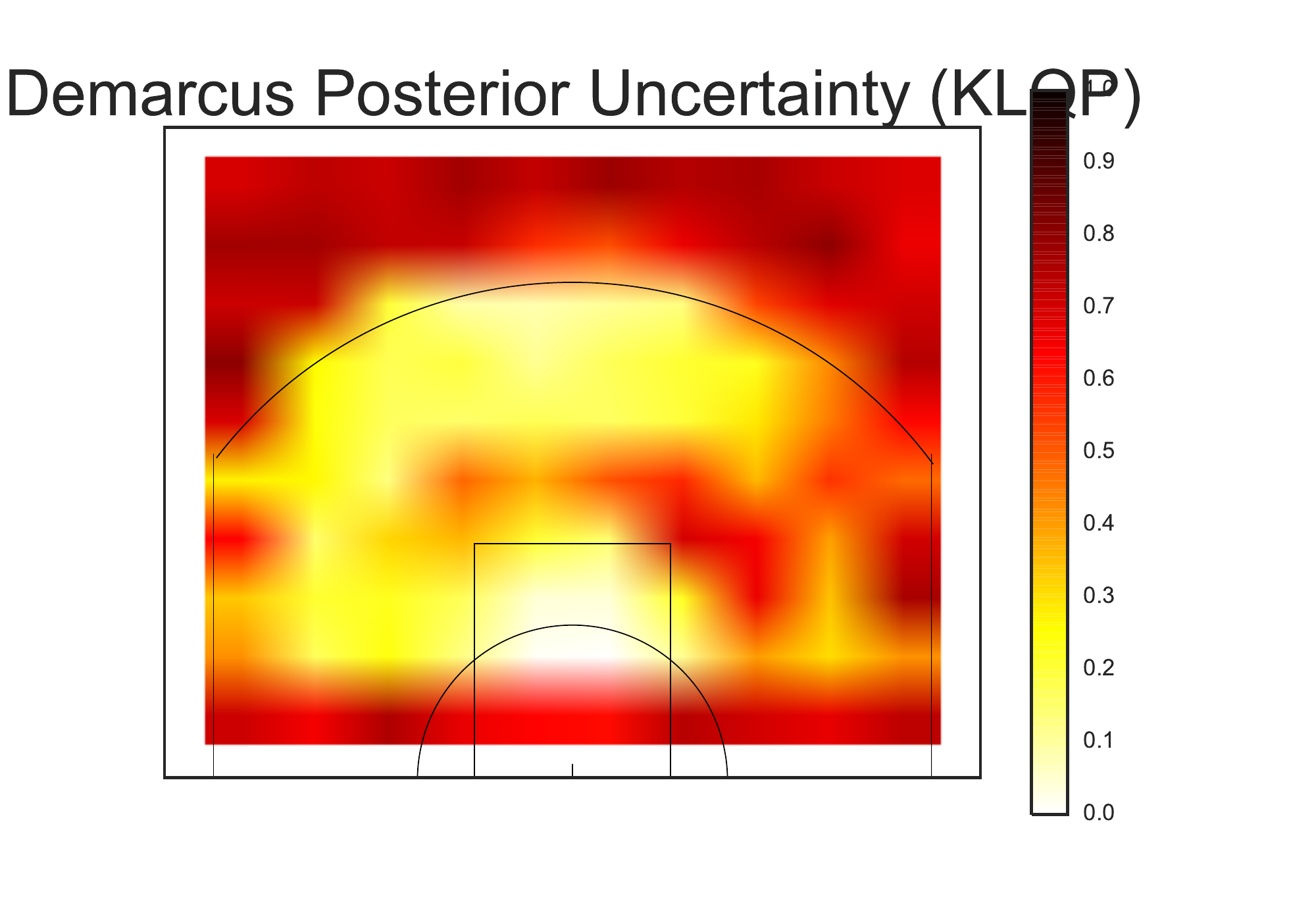}}
    {\includegraphics[scale=0.21]{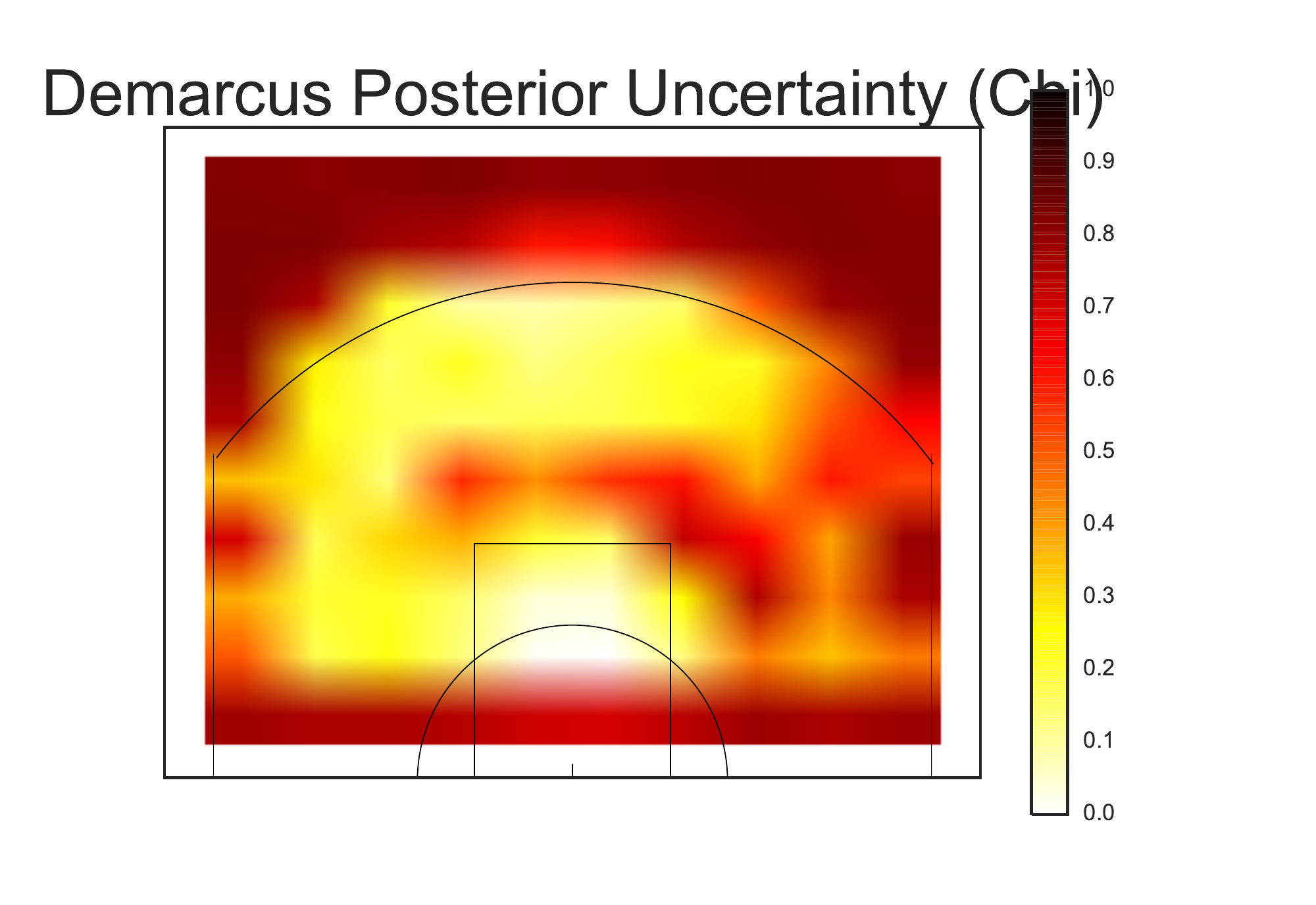}}
    {\includegraphics[scale=0.21]{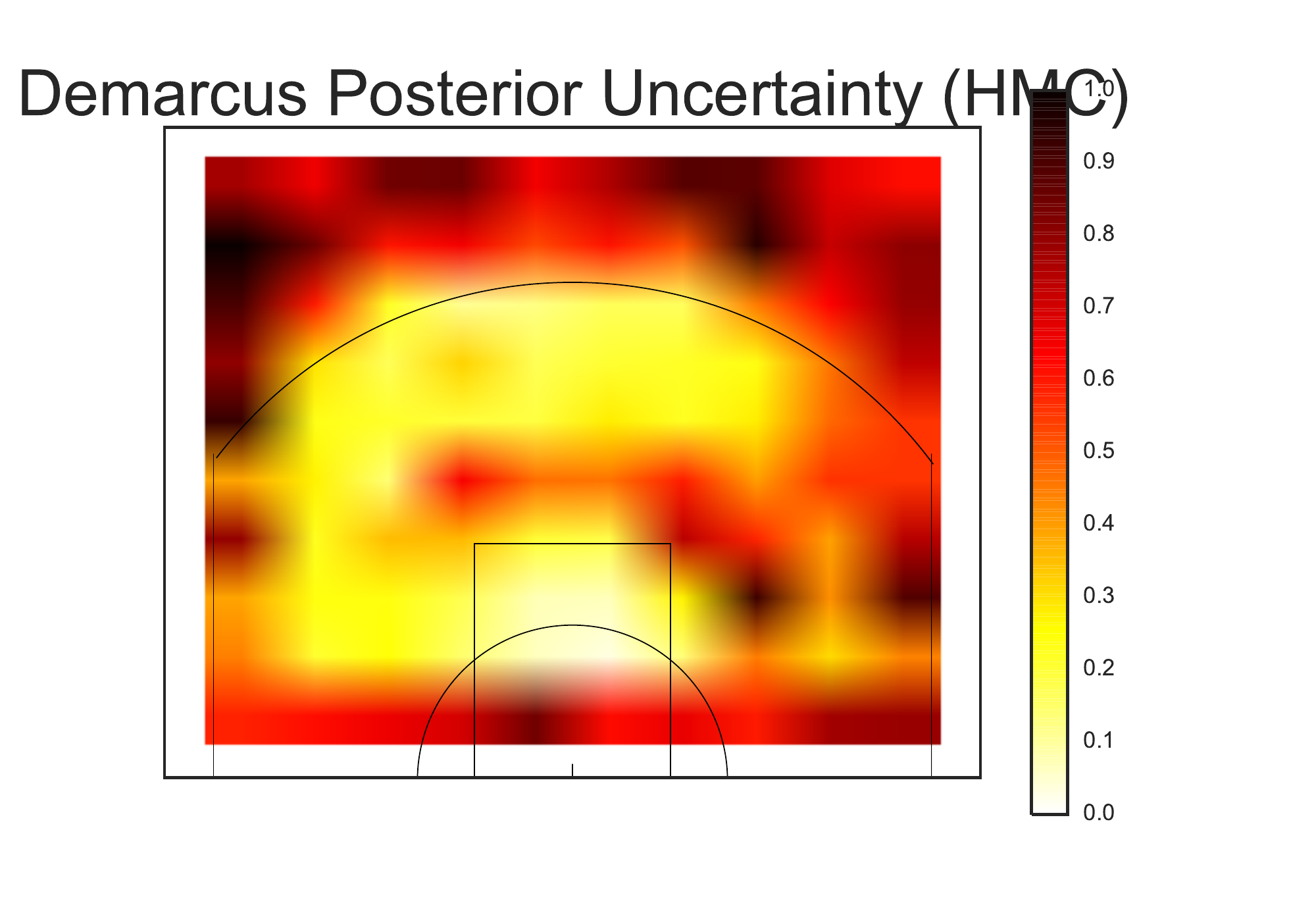}}
    \caption{Basketball players shooting profiles as inferred by
    \gls{BBVI} ~\citep{ranganath2014black}, \gls{chiVI} (this paper), 
    and \gls{HMC}. The top row displays the raw data, consisting
    of made shots (\green{green}) and missed shots (\red{red}). 
    The second and third rows display the posterior intensities inferred
    by \gls{BBVI}, \gls{chiVI}, and \gls{HMC} for Stephen
    Curry and Demarcus Cousins respectively. Both \gls{BBVI} and \gls{chiVI} capture the
    shooting behavior of both players in terms of the posterior mean. 
    The last two rows display the posterior uncertainty
    inferred by \gls{BBVI}, \gls{chiVI}, and \gls{HMC} for Stephen
    Curry and Demarcus Cousins respectively. \gls{chiVI} tends to get higher posterior uncertainty for
    both players in areas where data is scarce compared to \gls{BBVI}. This illustrates
    the variance underestimation problem of
    \gls{klVI}, which is not the case for \gls{chiVI}. 
    More player profiles with posterior mean and uncertainty 
    estimates can be found in the appendix.}
    \label{fig:players}
\end{figure*}

\gls{GP} classification is an alternative to probit
regression. The posterior is analytically intractable
because the likelihood is not conjugate to the prior. Moreover, the
posterior tends to be skewed. \gls{EP} has been the method of choice
for approximating the posterior~\citep{kuss2005assessing}.
We choose a factorized Gaussian for the variational
distribution $q$ and fit its mean and $\log$ variance parameters.

With UCI benchmark datasets, we compared the predictive
performance of \gls{chiVI} to \gls{EP} and Laplace.
Table \ref{table:error_rates} summarizes the results.
The error rates for \gls{chiVI} correspond to the 
average of $10$ error rates obtained by
dividing the data into $10$ folds, applying \gls{chiVI} 
to $9$ folds to learn the variational parameters 
and performing prediction on the remainder.
The kernel hyperparameters were chosen using grid search.
The error rates for the other methods correspond to the best results
reported in \cite{kuss2005assessing} and \cite{kim2003ep}.
On all the datasets \gls{chiVI} performs as 
well or better than \gls{EP} and Laplace.

\subsection{Cox Processes}
Finally we study Cox processes. They are Poisson processes
with stochastic rate functions. They capture dependence between the
frequency of points in different regions of a space.  We apply Cox
processes to model the spatial locations of shots (made and missed) from the
2015-2016 NBA season~\cite{miller2014factorized}.
The data are from $308$ NBA players who 
took more than $150,000$~shots in total.
The $n^{th}$ player's set of $M_n$ shot attempts are
$\mbx_n = \{\mbx_{n,1}, ..., \mbx_{n, M_n}\}$,
and the location of the $m^{th}$ shot by the $n^{th}$ player in the basketball court is
$\mbx_{n,m} \in [-25, 25]\times [0 , 40]$.
Let $\mathcal{PP}(\mblambda)$ denote a 
Poisson process with intensity function $\mblambda$, 
and $\mbK$ be a covariance matrix resulting from a kernel 
applied to every location of the court. 
The generative process for the $n^{th}$ player's shot is
\begin{align*}
\mbK_{i,j} &= k(\mbx_i, \mbx_j) = \sigma^2 \exp(-\frac{1}{2\phi^2} ||\mbx_i-\mbx_j||^2) \\
\textbf{f} &\sim \mathcal{GP}(0, k(\cdot, \cdot))
\text{ ; } \mblambda = \exp(\textbf{f}) \text{ ; }
\mbx_{n,k} \sim \mathcal{PP}(\mblambda) \text{ for } k \in \{1, ..., M_n \}.
\end{align*}
The kernel of the Gaussian process encodes the spatial 
correlation between different areas of the basketball court.
The model treats the $N$ players as independent.
But the kernel $\mbK$ introduces correlation 
between the shots attempted by a given player.

Our goal is to infer the intensity functions $\lambda(.)$ for each
player. We compare the shooting
profiles of different players using these inferred
intensity surfaces. The results are shown in Figure\nobreakspace \ref {fig:players}.
The shooting profiles of Demarcus Cousins and Stephen Curry are captured by
both \gls{BBVI} and \gls{chiVI}. \gls{BBVI} has lower
posterior uncertainty while \gls{chiVI} provides more overdispersed
solutions. We plot the profiles for two more players, LeBron James and Tim Duncan,
in the appendix.

\begin{table}[t]
\caption{Average $L_1$ error for posterior uncertainty estimates (ground truth from
\gls{HMC}). We find that \gls{chiVI} is similar to or better than \gls{BBVI} at
capturing posterior uncertainties. Demarcus Cousins, who plays center, stands out in particular. 
His shots are concentrated near the basket, so the posterior is 
uncertain over a large part of the court~Figure\nobreakspace \ref {fig:players}.\\}
\centering
\begin{tabular}{ccccc}
\toprule
 & Curry & Demarcus & Lebron & Duncan \\\midrule
\gls{chiVI} & \textbf{0.060} & \textbf{0.073}  &  0.0825 &   \textbf{0.0849} \\
\gls{BBVI} &  0.066 &  0.082 &   \textbf{0.0812} &  0.0871 \\\bottomrule
\end{tabular}
\label{tab:posterior-error}
\end{table}
In~Table\nobreakspace \ref {tab:posterior-error}, we compare the posterior uncertainty estimates
of \gls{chiVI} and \gls{BBVI} to that of \gls{HMC}, a computationally expensive 
Markov chain Monte Carlo procedure that we treat as exact. 
We use the average $L_1$ distance from \gls{HMC} as error measure.
We do this on four different players: Stephen Curry, Demarcus Cousins, LeBron James, and Tim Duncan. 
We find that \gls{chiVI} is similar or better than \gls{BBVI}, especially on players 
like Demarcus Cousins who shoot in a limited part of the court.
 
\section{Discussion}
\label{sec:discussion}
We described \gls{chiVI}, a black box algorithm that minimizes the
$\chi$-divergence by minimizing the \gls{CUBO}.
We motivated \gls{chiVI} as a useful alternative to \gls{EP}. 
We justified how the approach used in \gls{chiVI} enables 
upper bound minimization contrary to existing $\alpha$-divergence 
minimization techniques. This enables sandwich estimation 
using variational inference instead of \acrlong{MCMC}. 
We illustrated this by showing how to use \gls{chiVI} in concert with \gls{klVI} to 
sandwich-estimate the model evidence. 
Finally, we showed that \gls{chiVI} is an effective algorithm 
for Bayesian probit regression, Gaussian process 
classification, and Cox processes. 

Performing \gls{VI} via upper bound minimization, and hence enabling overdispersed 
posterior approximations, sandwich estimation, and model selection, comes with a cost. 
Exponentiating the original \gls{CUBO} bound leads to high variance during optimization even 
with reparameterization gradients. Developing variance reduction schemes for these types of 
objectives (expectations of likelihood ratios) is an open research problem; solutions 
will benefit this paper and related approaches. 

\section*{Acknowledgments}

We thank Alp Kucukelbir, Francisco J. R. Ruiz, 
Christian A. Naesseth, Scott W. Linderman, Maja Rudolph,  and Jaan Altosaar 
for their insightful comments. This work is supported by NSF IIS-1247664, 
ONR N00014-11-1-0651, DARPA PPAML FA8750-14-2-0009, 
DARPA SIMPLEX N66001-15-C-4032, the Alfred
P. Sloan Foundation, and the John Simon Guggenheim Foundation.

\bibliography{main}
\bibliographystyle{unsrt}

\appendix
\section{Proof of Sandwich Theorem}\label{ap:proof}
We denote by $\mbz$ the latent variable and $\mbx$ the data.
Assume $\mbz \in R^D$.

We first  show that \gls{CUBO}$_n$ is a nondecreasing 
function of the order $n$ of the $\chi$-divergence.
Denote by the triplet $(\Omega, \mathcal{F}, Q)$
the probability space induced by the variational distribution $q$
where $\Omega$ is a subspace of $R^D$,
$\mathcal{F}$ is the corresponding Borel sigma algebra,
and $Q$ is absolutely continuous with respect to 
the Lebesgue measure $\mu$ and is such that 
$dQ(\mbz) = q(\mbz)dz$.
Define $w = \frac{p(\mbx, \mbz)}{q(\mbz)}$. We can rewrite $\gls{CUBO}_n$ as:
\begin{align*}
	\gls{CUBO}_n 
	&= \frac{1}{n} \log E_q[w^n]
	=  \log \Big( (E_q[w^n])^{\frac{1}{n}} \Big) 
\end{align*}
Since $\log$ is nondecreasing, it is enough to 
show\\ $n \mapsto (E_q[w^n])^{\frac{1}{n}} $ is nondecreasing.
This function is the $L_n$ norm in the space defined above:
\begin{align*}
	(E_q[w^n])^{\frac{1}{n}} 
	&= \Big (\int_{\Omega}^{} |w|^n dQ \Big)^{\frac{1}{n}}
	= \Big ( \int_{\Omega}^{} |w|^n q(\mbz)d\mbz\Big)^{\frac{1}{n}}
\end{align*}
This is a nondecreasing function of $n$ by virtue of
the Lyapunov inequality.

We now show the second claim in the sandwich theorem, 
namely that the limit when $n \rightarrow 0$ of 
$\gls{CUBO}_n$ is the \gls{ELBO}.
Since $\gls{CUBO}_n$ is a monotonic function of $n$
and is bounded from below by \gls{ELBO}, it admits a limit
when $n \rightarrow 0$. Call this limit $L$. We show 
$L = \gls{ELBO}$. 
On the one hand, since $\gls{CUBO}_n \geq \gls{ELBO}$
for all $n > 0$, we have $L \geq \gls{ELBO}$.
On the other hand, since $\log t \leq t - 1$;  $\forall t > 0$ 
we have 
\begin{align*}
	\gls{CUBO}_n 
	&= \frac{1}{n} \log E_q[w^n]
	\leq \frac{1}{n} \Big [ E_q[w^n] -1  \Big] = E_q\Big [ \frac{w^n -1}{n} \Big]
\end{align*}
$f: n \mapsto w^n$ is differentiable and furthermore\\ 
$f'(0) = \lim_{n\to0} \Big [ \frac{w^n -1}{n} \Big] = \log w$.
Therefore $\exists n_0 > 0$ such that 
$|\frac{w^n - 1}{n} - \log w| < 1$ $\forall n < n_0$.\\
Since $||\frac{w^n - 1}{n}| - \log w|  <  |\frac{w^n - 1}{n} - \log w|$, 
we have \\ $|\frac{w^n - 1}{n}| < 1 + \log w$ which is $E_q$-integrable.
Therefore by Lebesgue's dominated convergence theorem:
$ \lim_{n\to0} E_q\Big [ \frac{w^n -1}{n} \Big] 
= E_q \Big [\lim_{n\to0} \frac{w^n -1}{n} \Big]
= E_q [\log w] = \gls{ELBO}$.
Since $\gls{CUBO}_n$ converges when $n \rightarrow 0$
and $\gls{CUBO}_n  \leq  E_q\Big [ \frac{w^n -1}{n} \Big]$ $\forall n$,
we establish $L \leq  \lim_{n\to0} E_q\Big [ \frac{w^n -1}{n} \Big] = \gls{ELBO}$.
The conclusion follows.

\section{The \gls{chiVI} algorithm for small datasets}
In the main text we derived a subsampling version of \gls{chiVI}. For very small 
datasets, the average likelihood technique is not needed. The algorithm 
then uses all the data at each iteration and is summarized in~Algorithm\nobreakspace \ref {alg:chivi}. 
\begin{algorithm}[!hbpt]
  \caption{\gls{chiVI} without average likelihoods}
  \SetAlgoLined
  \DontPrintSemicolon
  \BlankLine
  \KwIn{Data $\mbx$, Model $p(\mbx,\mbz)$, Variational family $q(\mbz;
  \mblambda)$.}
  \BlankLine
  \textbf{Output}: Variational parameters $\mblambda$.\;
  \BlankLine
  Initialize $\mblambda$ randomly.
  \BlankLine
  \While{\textnormal{not converged}}{
    \BlankLine
    Draw $S$ samples $\mbz^{(1)},...,\mbz^{(S)}$ from $q(\mbz; \mblambda)$.
    \BlankLine
    Set $\rho_t$ from a Robbins-Monro sequence.
    \BlankLine
    Set $\log \mbw^{(s)} = \log p(\mbx,\mbz^{(s)}) - \log
    q(\mbz^{(s)}; \mblambda_t)$, $s\in \{1, ..., S\}$.
    \BlankLine
    Set $c = \displaystyle\max_{s} \log \mbw^{(s)} $.
    \BlankLine
    Set $\mbw^{(s)} =\exp( \log \mbw^{(s)}  - c), s\in \{1, ..., S\}$.
    \BlankLine
    Update $\mblambda_{t+1} = \mblambda_{t} - \frac{(1-n)\cdot \rho_t}{S}
    \sum_{s =
    1}^{S}\Big[\Big(\mbw^{(s)}\Big)^n
                            \nabla_{\mblambda}\log q(\mbz^{(s)}; \mblambda_t)\Big]$.
  }
  \label{alg:chivi}
\end{algorithm}

\section{Approximately minimizing an $f$-divergence with \gls{chiVI}}\label{ap:fdiv}
In this section we provide a proof that minimizing an
$f$-divergence can be done by minimizing a sum of
$\chi$-divergencesThese individual $\chi$-divergences 
can then be optimized via \gls{chiVI}. Consider
\begin{equation*}
D_f(p\gg q) = \int_{}^{} f\Big(\frac{p(x)}{q(x)}\Big) q(x) dx
\end{equation*}
Without loss of generality assume $f$ is analytic.
The Taylor expansion of $f$ around a given point $x_0$ is
\begin{equation*}
	f(x) = f(x_0) + f'(x_0)(x - x_0) + \sum_{i = 2}^{\infty} f^{(i)}(x_0)\frac{(x-x_0)^i}{i!}
\end{equation*}
Therefore
\begin{align*}
  D_f(p\gg q) &= f(x_0) + f'(x_0)\Big(\mathbb{E}_{q(\mbz\g\mblambda)}\Big[\frac{p(x)}{q(x)}\Big] - x_0\Big) 
  		    + \mathbb{E}_{q(\mbz\g\mblambda)}\Big[\sum_{i = 2}^{\infty} \frac{f^{(i)}(x_0)}{i!}\Big(\frac{p(x)}{q(x)}-x_0\Big)^i\Big] \\
 		    &= f(x_0) + f'(x_0)(1-x_0) 
		    + \sum_{i = 2}^{\infty} \frac{f^{(i)}(1)}{i!} \mathbb{E}_{q(\mbz\g\mblambda)}\Big[\Big(\frac{p(x)}{q(x)}-1\Big)^i \Big]
\end{align*}
where we switch summation and expectation by invoking Fubini's theorem.
In particular if we take $x_0=1$ the linear terms are zero and we end up with:
\begin{align*}
 D_f(p\gg q) &= \sum_{i = 2}^{\infty} \frac{f^{(i)}(1)}{i!} \mathbb{E}_{q(\mbz\g\mblambda)}\Big[\Big(\frac{p(x)}{q(x)}-1\Big)^i \Big] 
 		   = \sum_{i = 2}^{\infty} \frac{f^{(i)}(1)}{i!}  D_{\chi ^i}(p\gg q)
\end{align*}
If $f$ is not analytic but $k$ times differentiable for some
$k$ then the proof still holds considering the Taylor expansion of $f$ up to the order $k$.

\section{Importance sampling}\label{ap:is}
In this section we establish the relationship
between $\chi^2$-divergence minimization and importance sampling.
Consider estimating the marginal likelihood $I$ with importance sampling:
\begin{align*}
  I &= p(\mbx)
  = \int_{}^{} p(\mbx, \mbz) d\mbz 
  = \int_{}^{} \frac{p(\mbx, \mbz)}{q(\mbz; \mblambda)} q(\mbz; \mblambda) d\mbz
  = \int_{}^{} w(\mbz) q(\mbz; \mblambda) d\mbz
\end{align*}
The Monte Carlo estimate of $I$ is
\begin{equation*}
\hat{I} = \frac{1}{B} \sum_{b=1}^{B} w(\mbz^{(b)})
\end{equation*}
where $\mbz^{(1)}, ... ,\mbz^{(B)}  \sim q(\mbz; \mblambda)$.
The variance of $\hat{I}$ is
\begin{align*}
  \operatorname{Var}(\hat{I}) 
  	&=  \frac{1}{B} [\mathbb{E}_{q(\mbz; \mblambda)}
  		(w(\mbz^{(b)})^2) - (\mathbb{E}_{q(\mbz; \mblambda)}(w(\mbz^{(b)})))^2] 
          = \frac{1}{B} \Big[\mathbb{E}_{q(\mbz; \mblambda)}
         	 \Big(\Big(\frac{p(\mbx, \mbz^{(1)})}{q(\mbz^{(1)}; \mblambda)}\Big)^2\Big) - p(\mbx)^2\Big]
\end{align*}
Therefore minimizing this variance is equivalent to minimizing the quantity
\begin{equation*}
 \mathbb{E}_{q(\mbz; \mblambda)}\Big(\Big(\frac{p(\mbx, \mbz^{(1)})}{q(\mbz^{(1)}; \mblambda)}\Big)^2\Big)
\end{equation*}
which is equivalent to minimizing the  $\chi ^2$-divergence.

\section{General properties of the $\chi$-divergence}\label{ap:properties}
In this section we outline several properties of the $\chi$-divergence. \\

\parhead{Conjugate symmetry} Define
\begin{equation*}
 f^*(u) = uf(\frac{1}{u})
\end{equation*}
to be the conjugate of $f$. $f^*$ is also convex and satisfies $f^*(1) = 0$.
Therefore $D_f^*(p\gg q) $ is a valid divergence in the $f$-divergence family and:
\begin{align*}
D_f(q\gg p) &= \int_{}^{} f\Big(\frac{q(x)}{p(x)}\Big) p(x) dx
= \int_{}^{} \frac{q(x)}{p(x)} f^*\Big(\frac{p(x)}{q(x)}\Big) p(x) dx
= D_{f^*}(p\gg q)
\end{align*}
$D_f(q\gg p)$ is symmetric if and only if $f = f^*$ which is not the case here.
To symmetrize the divergence one can use
\begin{equation*}
 D(p \gg q)  = D_f(p\gg q) + D_f^*(p\gg q)
\end{equation*}

\parhead{Invariance under parameter transformation.}
Let $y = u(x)$ for some function $u$.
Then by Jacobi $p(x)dx = p(y)dy$ and $q(x)dx = q(y)dy$.
\begin{align*}
D_{\chi ^n}(p(x)\gg q(x)) &= \int_{x_0}^{x_1} \Big(\frac{p(x)}{q(x)}\Big)^n q(x) dx - 1 
=  \int_{y_0}^{y_1} \Big(\frac{p(y) \frac{dy}{dx}}{q(y) \frac{dy}{dx}}\Big)^n q(y)dy - 1 \\
&= \int_{y_0}^{y_1} \Big(\frac{p(y)}{q(y)}\Big)^n q(y)dy - 1 
= D_{\chi ^n}(p(y)\gg q(y))
\end{align*}

\parhead{Factorization for independent distributions.}
Consider taking $p(x,y) = p_1(x)p_2(y)$ and $q(x,y) = q_1(x)q_2(y)$.
\begin{align*}
D_{\chi ^n}(p(x,y)\gg q(x,y)) &= \int_{}^{} \frac{p(x,y)^n}{q(x,y)^{n-1}} dx dy 
= \int_{}^{} \frac{p_1(x)^np_2(y)^n}{q_1(x)^{n-1}q_2(y)^{n-1}} dx dy  \\
&= \Big(\int_{}^{} \frac{p_1(x)^n}{q_1(x)^{n-1}} dx \Big) \cdot 
 \Big(\int_{}^{} \frac{p_2(y)^n}{q_2(y)^{n-1}} dy \Big) \\
&= D_{\chi ^n}(p_1(x)\gg q_1(x)) \cdot 
D_{\chi ^n}(p_2(y)\gg q_2(y))
\end{align*}
Therefore $\chi$-divergence is multiplicative under 
independent distributions while KL is additive.

\parhead{Other properties.}
The $\chi$-divergence enjoys some other properties that 
it shares with all members of the $f$-divergence family
namely monotonicity with respect to the distributions and joint convexity.

\section{Derivation of the $\glsunset{CUBO}$ $\gls{CUBO}_n$}\label{ap:bound}
In this section we outline the derivation of $\gls{CUBO}_n$,
the upper bound to the marginal likelihood induced by 
the minimization of the $\chi$-divergence.\\
By definition:
\begin{align*}
  D_{\chi^n}(p(\mbz \g \mbx)\gg q(\mbz; \mblambda)) 
  	&= \mathbb{E}_{q(\mbz; \mblambda)}\Big[\Big(\frac{p(\mbz|\mbx)}{q(\mbz; \mblambda)}\Big)^n - 1 \Big]
\end{align*}
Following the derivation of \gls{ELBO}, we seek an
expression of $\log(p(\mbx))$ involving this divergence. 
We achieve that as follows:
\begin{align*}
\mathbb{E}_{q(\mbz; \mblambda)}\Big[\Big(\frac{p(\mbz \g \mbx)}{q(\mbz; \mblambda)}\Big)^n \Big]   
&=  1 + D_{\chi ^n}(p(\mbz \g \mbx)\gg q(\mbz; \mblambda)) 
 \mathbb{E}_{q(\mbz; \mblambda)}\Big[\Big(\frac{p(\mbx,\mbz)}{q(\mbz; \mblambda)}\Big)^n \Big]   \\
 &=  p(\mbx)^n[1 + D_{\chi ^n}(p(\mbz \g \mbx)\gg q(\mbz; \mblambda))]
 \end{align*}
 This gives the relationship
 \begin{align*}
  \log p(\mbx) &= \frac{1}{n}\log\mathbb{E}_{q(\mbz; \mblambda)}
  				\Big[\Big(\frac{p(\mbx,\mbz)}{q(\mbz; \mblambda)}\Big)^n \Big] -
  		     \frac{1}{n} \log(1 + D_{\chi ^n}(p(\mbz \g \mbx)\gg q(\mbz; \mblambda)))\\
 \log p(\mbx) &= \gls{CUBO}_n - \frac{1}{n} \log(1 + D_{\chi ^n}(p(\mbz|\mbx)\gg q(\mbz; \mblambda)))
\end{align*}
By nonnegativity of the divergence this last equation establishes the upper bound:
\begin{align*}
	\log p(\mbx) \leq \gls{CUBO}_n
\end{align*}

\section{Black Box Inference}\label{ap:grad}
In this section we derive the score gradient and the reparameterization
gradient for doing black box inference with the $\chi$-divergence.
\begin{equation*}
\gls{CUBO}_n(\mblambda) =
	\frac{1}{n}\log\mathbb{E}_{q(\mbz; \mblambda)}\Big[\Big(\frac{p(\mbx,\mbz)}{q(\mbz; \mblambda)}\Big)^n \Big]
\end{equation*}
where $\mblambda$ is the set of variational parameters. To minimize
$\gls{CUBO}_n(\mblambda)$ with respect to $\mblambda$ we need to resort to Monte Carlo.
To minimize $\gls{CUBO}_n(\mblambda)$ we consider the equivalent
minimization of $\exp\{n\cdot \gls{CUBO}(\mblambda)\}$. This enables unbiased estimation
of the noisy gradient used to perform black box inference with the $\chi$-divergence.

\parhead{The score gradient}
The score gradient of our objective function
\begin{align*}
\mbL = \exp\{n\cdot \gls{CUBO}(\mblambda)\}
\end{align*}
is derived below:
\begin{align*}
\nabla_{\mblambda} \mbL
&= \nabla_{\mblambda} \int_{}^{} p(\mbx,\mbz)^n q(\mbz; \mblambda)^{1-n} d\mbz 
= \int_{}^{} p(\mbx,\mbz)^n  \nabla_{\mblambda} q(\mbz; \mblambda)^{1-n} d\mbz \\
&=  \int_{}^{} p(\mbx,\mbz)^n (1-n) q(\mbz; \mblambda)^{-n}\nabla_{\mblambda}q(\mbz; \mblambda) d\mbz 
= (1-n)  \int_{}^{} (\frac{p(\mbx,\mbz)}{q(\mbz; \mblambda)})^n
 \nabla_{\mblambda}q(\mbz; \mblambda) d\mbz \\
&=  (1-n)  \int_{}^{} (\frac{p(\mbx,\mbz)}{q(\mbz; \mblambda)})^n
\nabla_{\mblambda}\log q(\mbz; \mblambda) q(\mbz; \mblambda) d\mbz 
= (1-n) \mathbb{E}_{q(\mbz; \mblambda)}\Big[\Big(\frac{p(\mbx,\mbz)}
{q(\mbz; \mblambda)}\Big)^n \nabla_{\mblambda}\log q(\mbz; \mblambda) \Big]
\end{align*}
where we switched differentiation and integration by invoking Lebesgue's dominated convergence theorem.
We estimate this gradient with the unbiased estimator:\begin{equation*}
\frac{(1-n)}{B} \sum_{b = 1}^{B}\Big[\Big(\frac{p(\mbx,\mbz^{(b)})}{q(\mbz^{(b)}; \mblambda)}\Big)^n
\nabla_{\mblambda}\log q(\mbz^{(b)}; \mblambda)\Big]
\end{equation*}

\parhead{Reparameterization gradient}
The reparameterization gradient empirically has lower variance
than the score gradient. We used it in our experiments.
Denote by $L$ the quantity $ \exp\{n\cdot \gls{CUBO}\}$.
Assume $\mbz = g(\mblambda, \epsilon)$ where $\epsilon \sim p(\epsilon)$.
Then
\begin{equation*}
 \hat{L} = \frac{1}{B} \sum_{b=1}^{B}  \Big(\frac{p(\mbx, g(\mblambda, \epsilon^{(b)}))}
 {q(g(\mblambda, \epsilon^{(b)}) ; \mblambda)}\Big)^n
\end{equation*}
is an unbiased estimator of $L$ and its gradient is given by
\begin{align*}
 \nabla_{\mblambda}  \hat{L}
     &= \frac{n}{B} \sum_{b=1}^{B} \Big(\frac{p(\mbx, g(\mblambda, \epsilon^{(b)}))}
 	{q(g(\mblambda, \epsilon^{(b)}) ; \mblambda)}\Big)^n
\nabla_{\mblambda}\log\Big(\frac{p(\mbx, g(\mblambda, \epsilon^{(b)}))}
{q(g(\mblambda, \epsilon^{(b)}) ; \mblambda)} \Big)
.
\end{align*}

\section{More illustrations}
\label{ap:simulations}
The following figures are results of various experimentations with the \gls{CUBO}.

\begin{figure}[!hbpt]
\centering
      \includegraphics[width=0.45\linewidth]{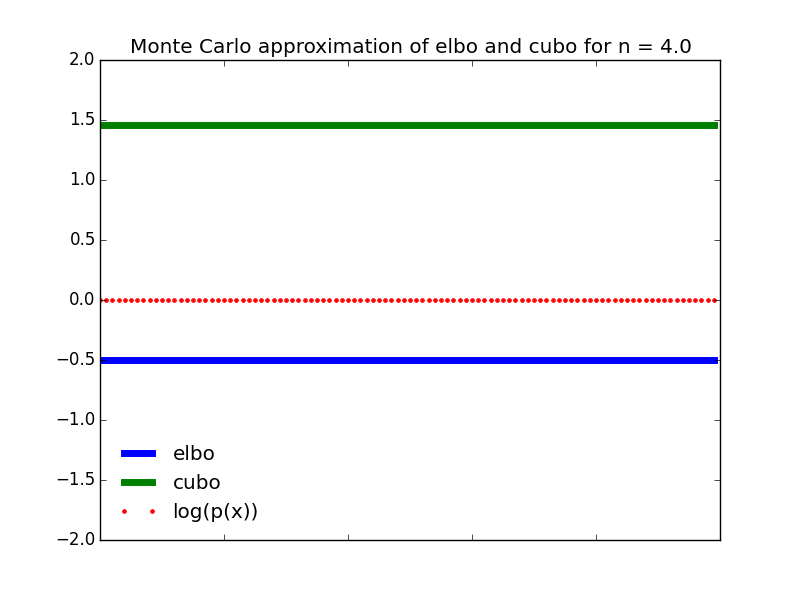}
   \includegraphics[width=0.45\linewidth]{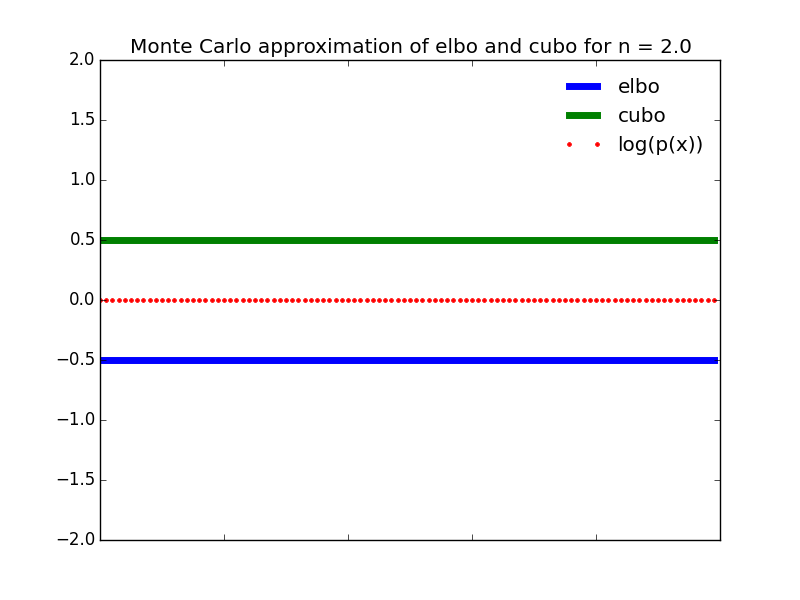}\\
    \includegraphics[width=0.45\linewidth]{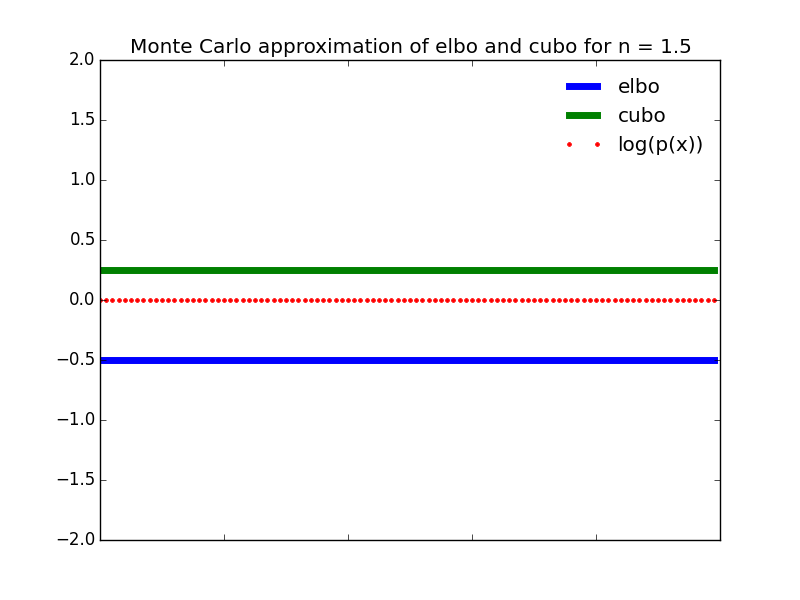}
    \includegraphics[width=0.43\linewidth]{img/sandwich.png}\\
    \includegraphics[width=0.45\linewidth]{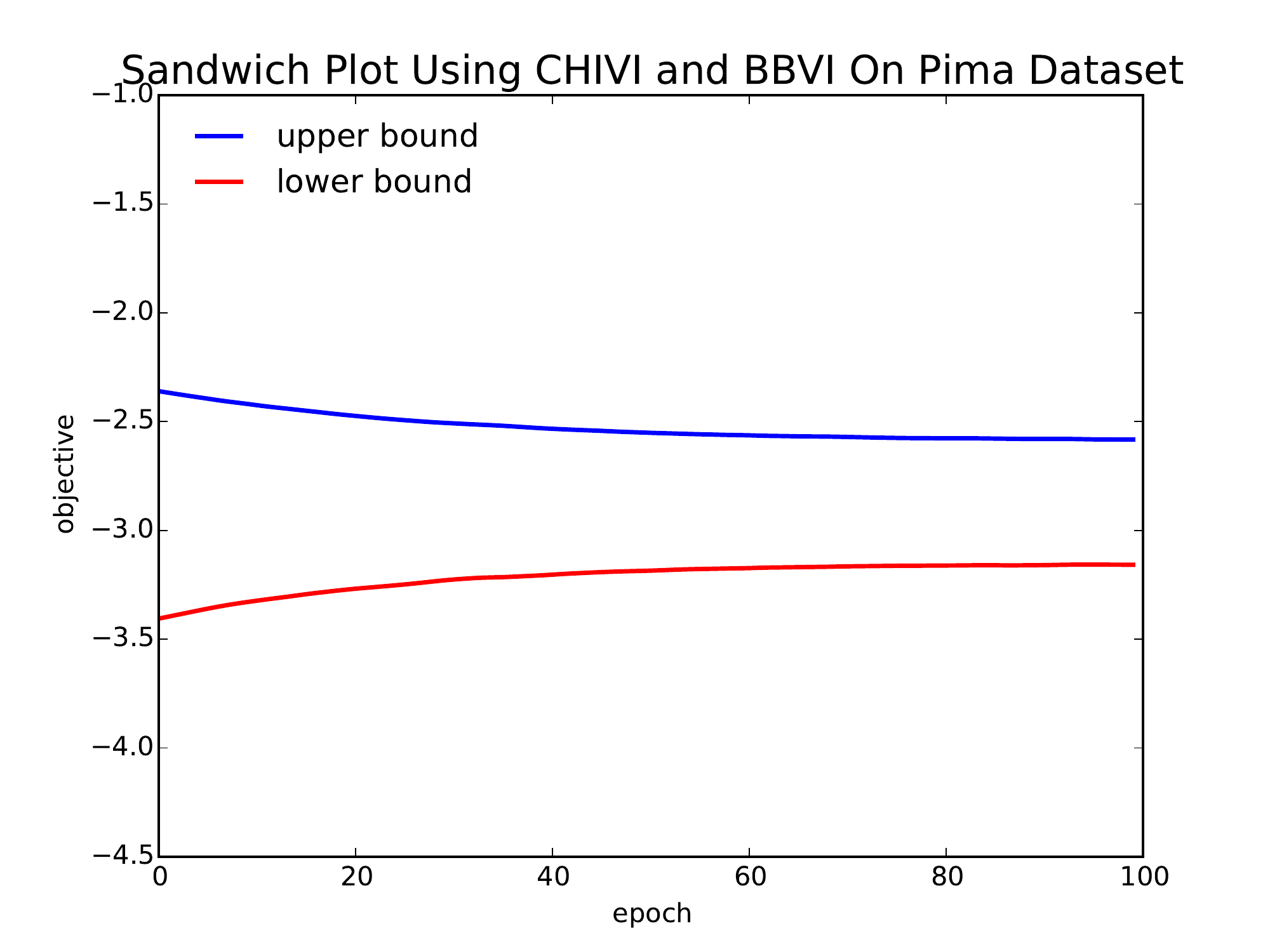}
    \includegraphics[width=0.45\linewidth]{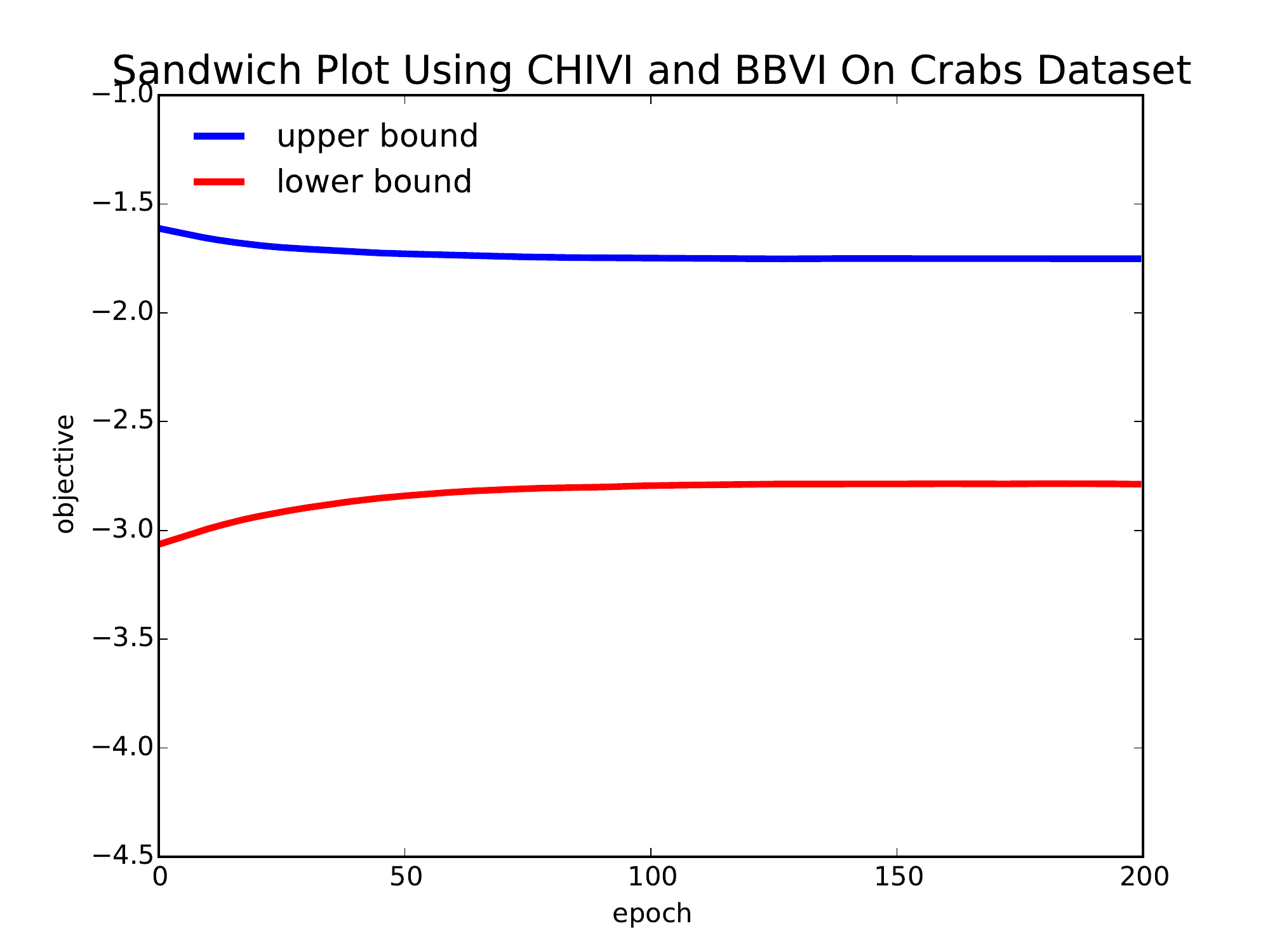}
   \caption{}
   \label{fig:cubovselbo1}
\caption{More sandwich plots via \gls{chiVI} and \gls{BBVI}. 
The first three plots show simulated sandwich gaps when the order of 
the $\chi$-divergence is $n = 4$, $n = 2$, and $n=1.5$ respectively. 
As we demonstrated theoretically, the gap closes as $n$ decreases. 
The fourth plot is a sandwich on synthetic data where we 
know the log marginal likelihood of the data. Here the gap tightens 
after only $100$ iterations. The final two plots 
are sandwiches on real UCI datasets.}
\end{figure}

\begin{figure*}[t]
   \centering
    {\includegraphics[scale=0.21]{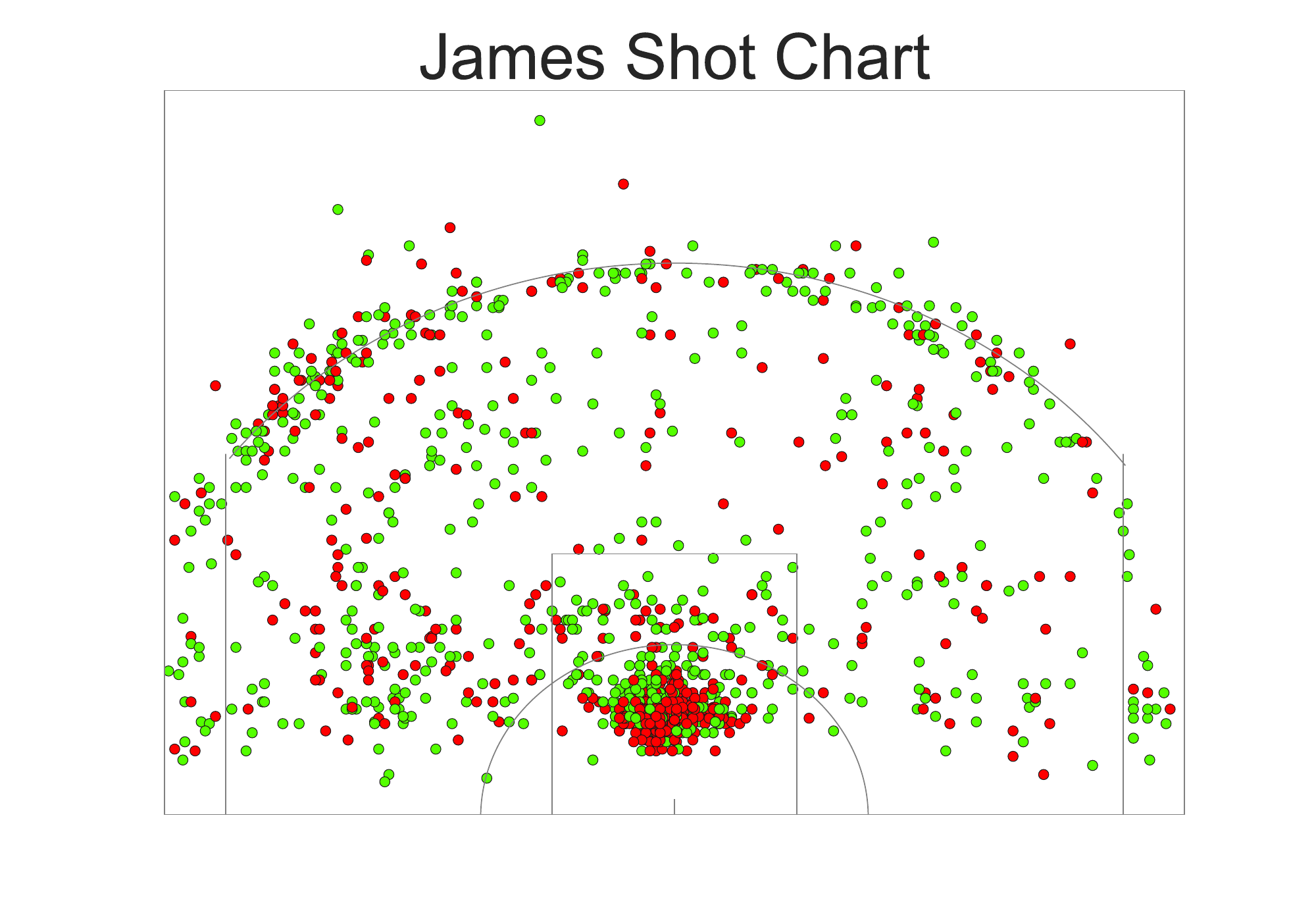}}
    {\includegraphics[scale=0.21]{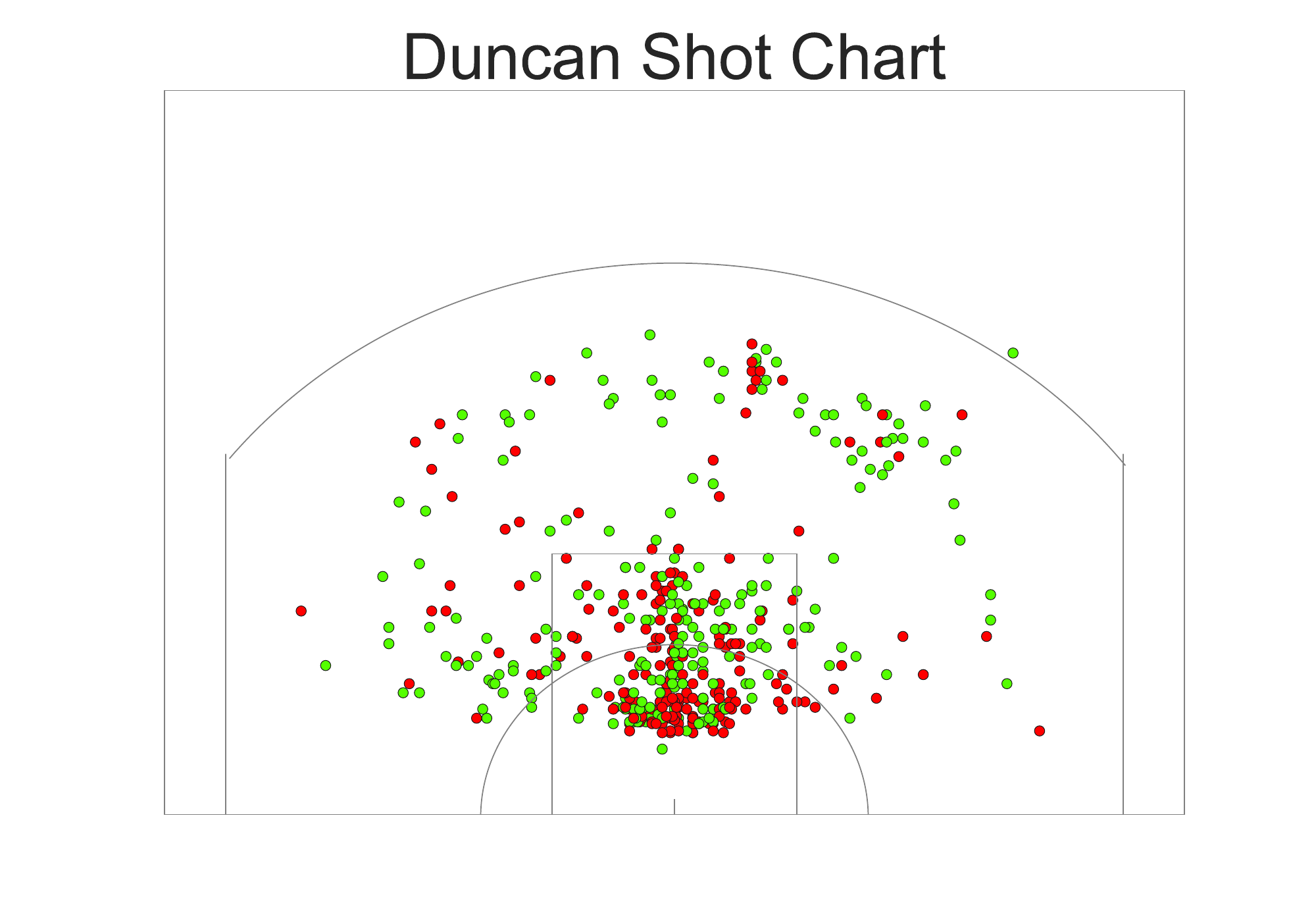}} \\
    {\includegraphics[scale=0.22]{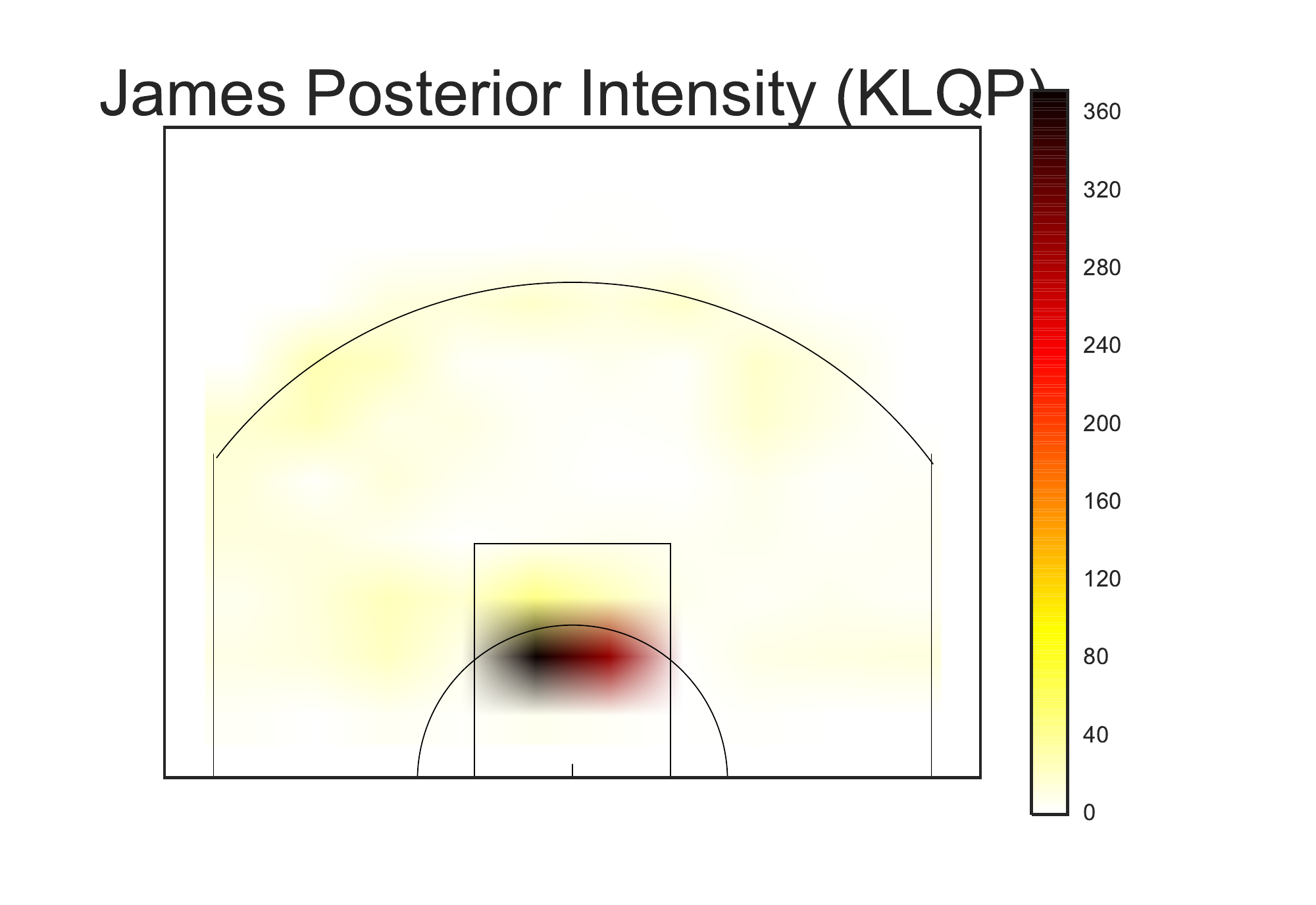}}
    {\includegraphics[scale=0.22]{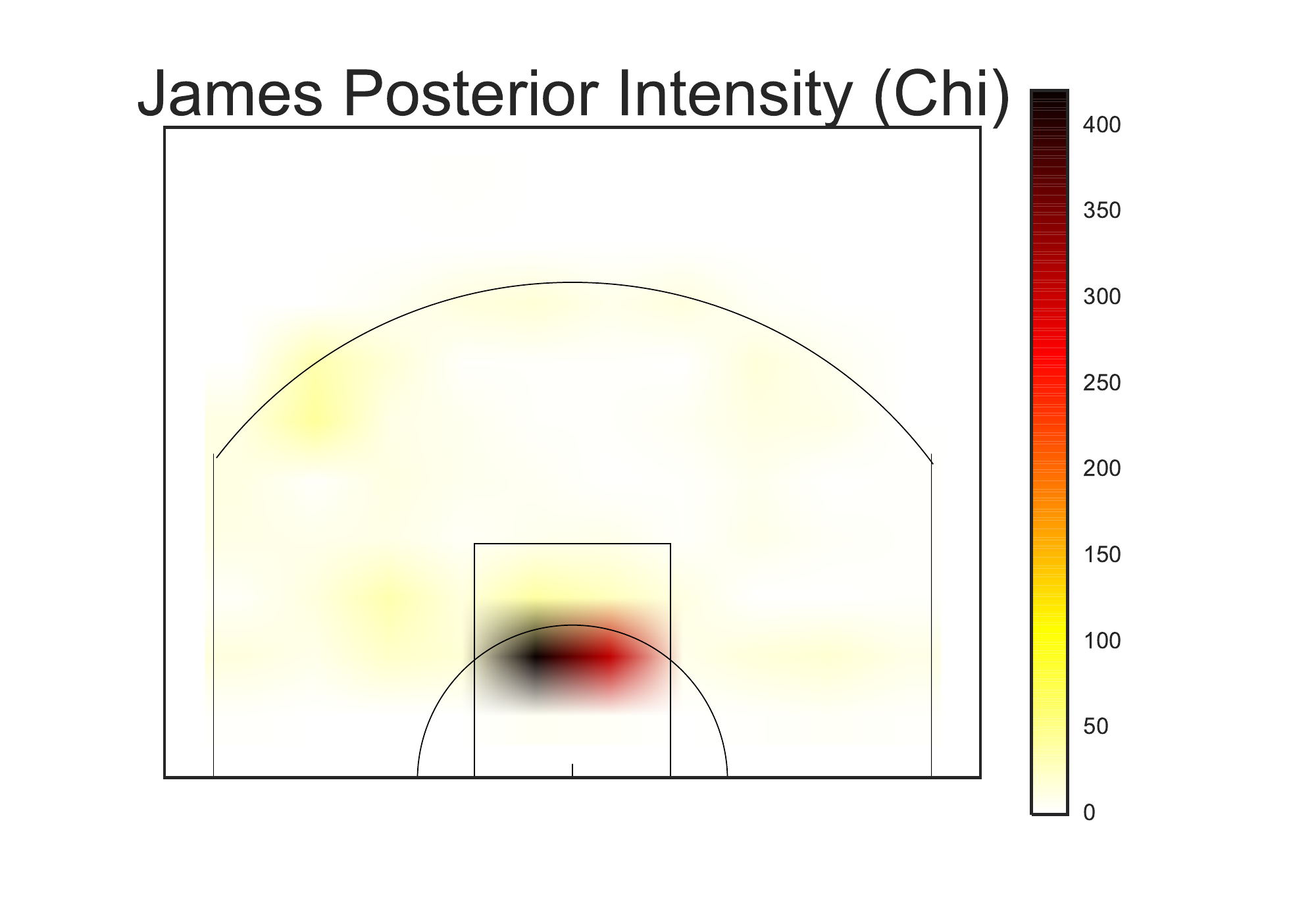}}
        {\includegraphics[scale=0.22]{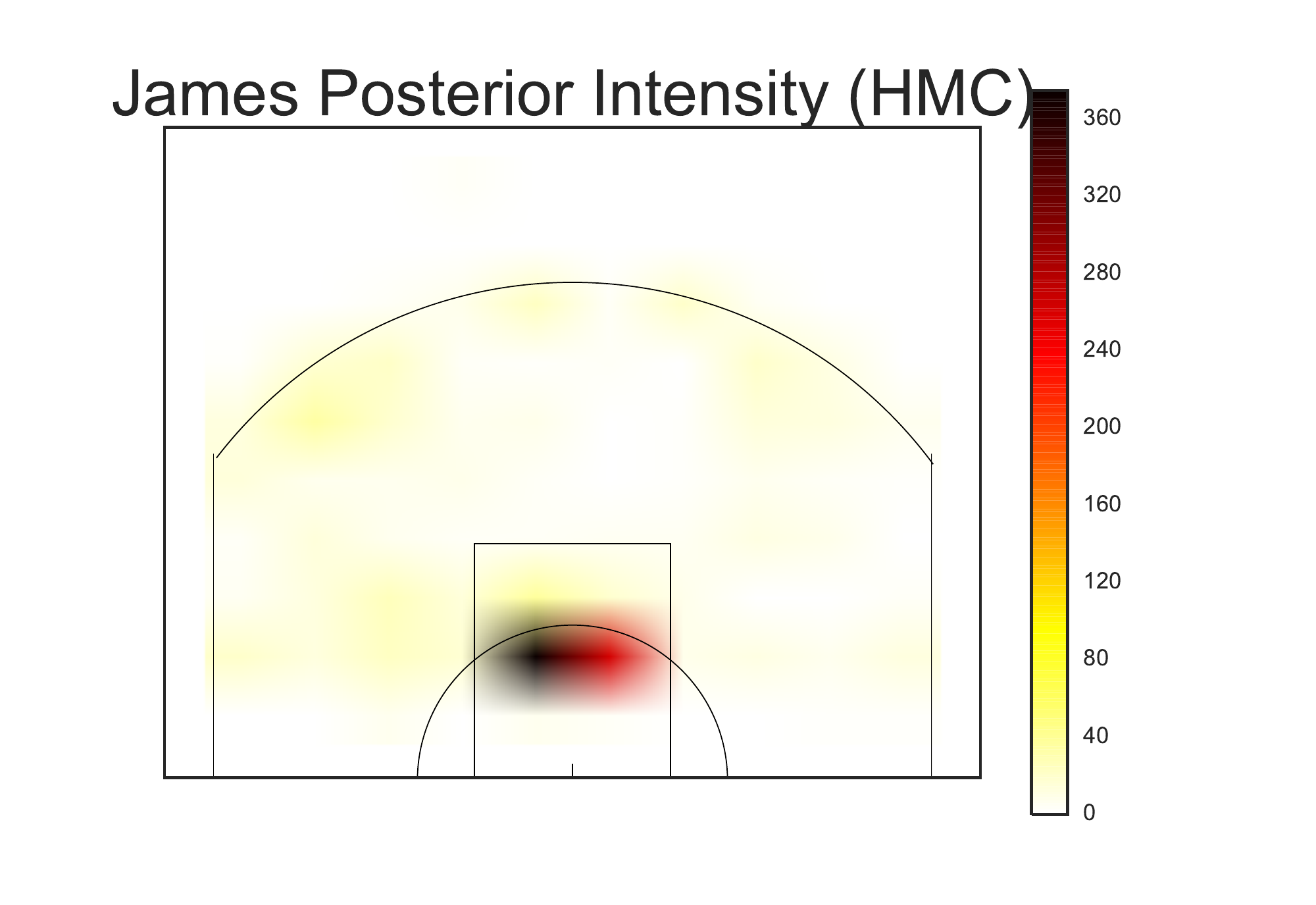}}\\
        {\includegraphics[scale=0.22]{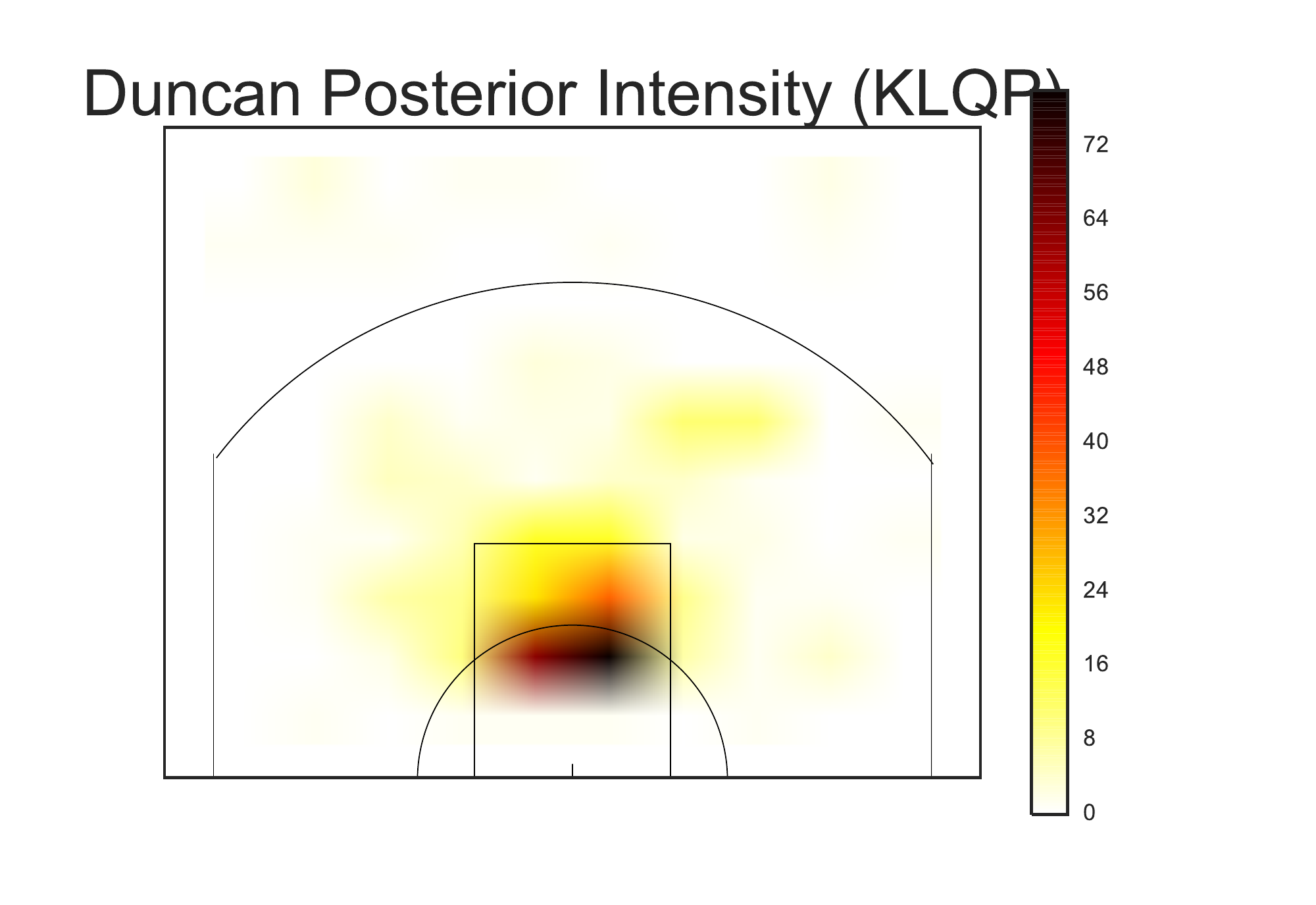}}
    {\includegraphics[scale=0.22]{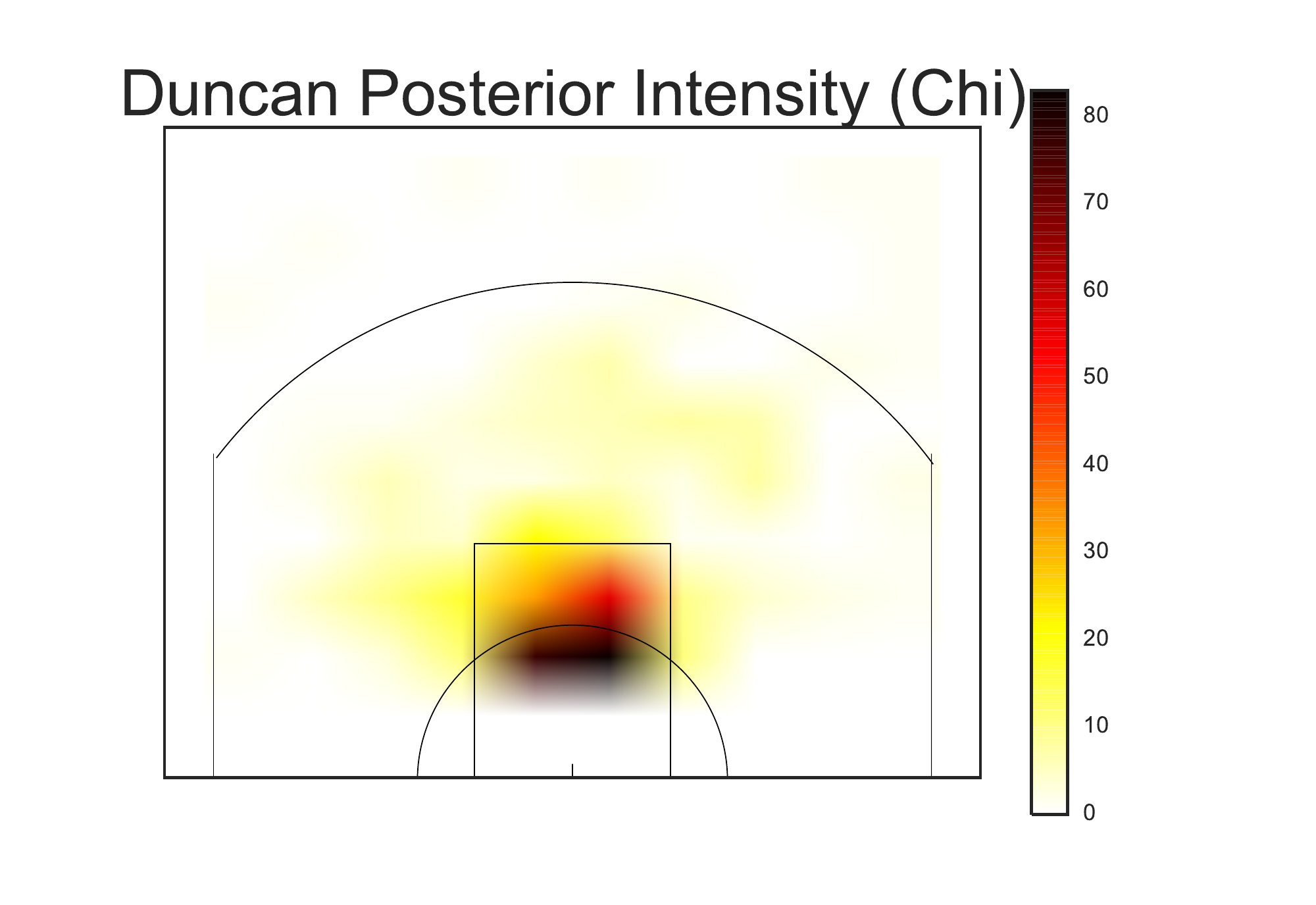}}
  {\includegraphics[scale=0.22]{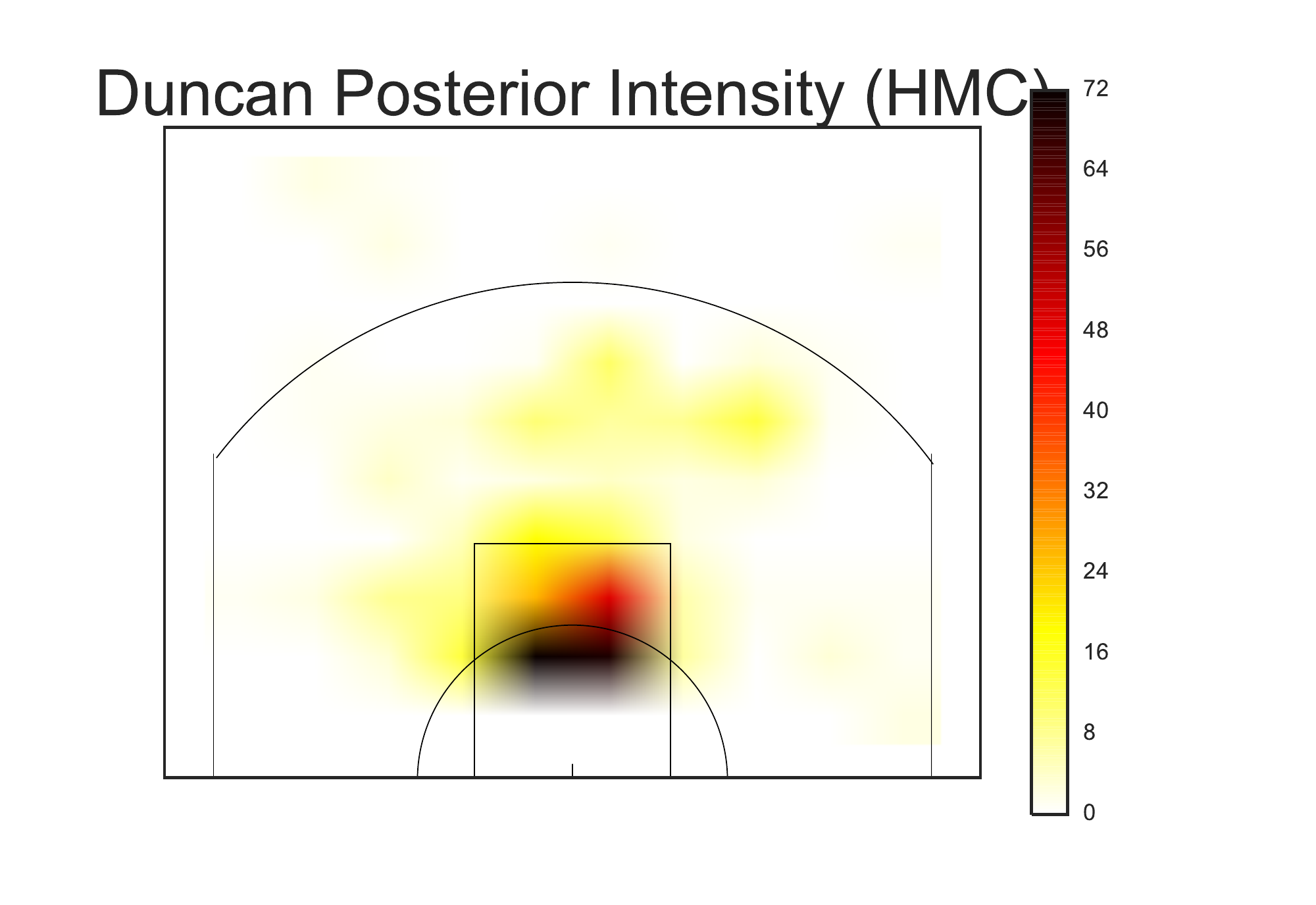}}\\
     {\includegraphics[scale=0.22]{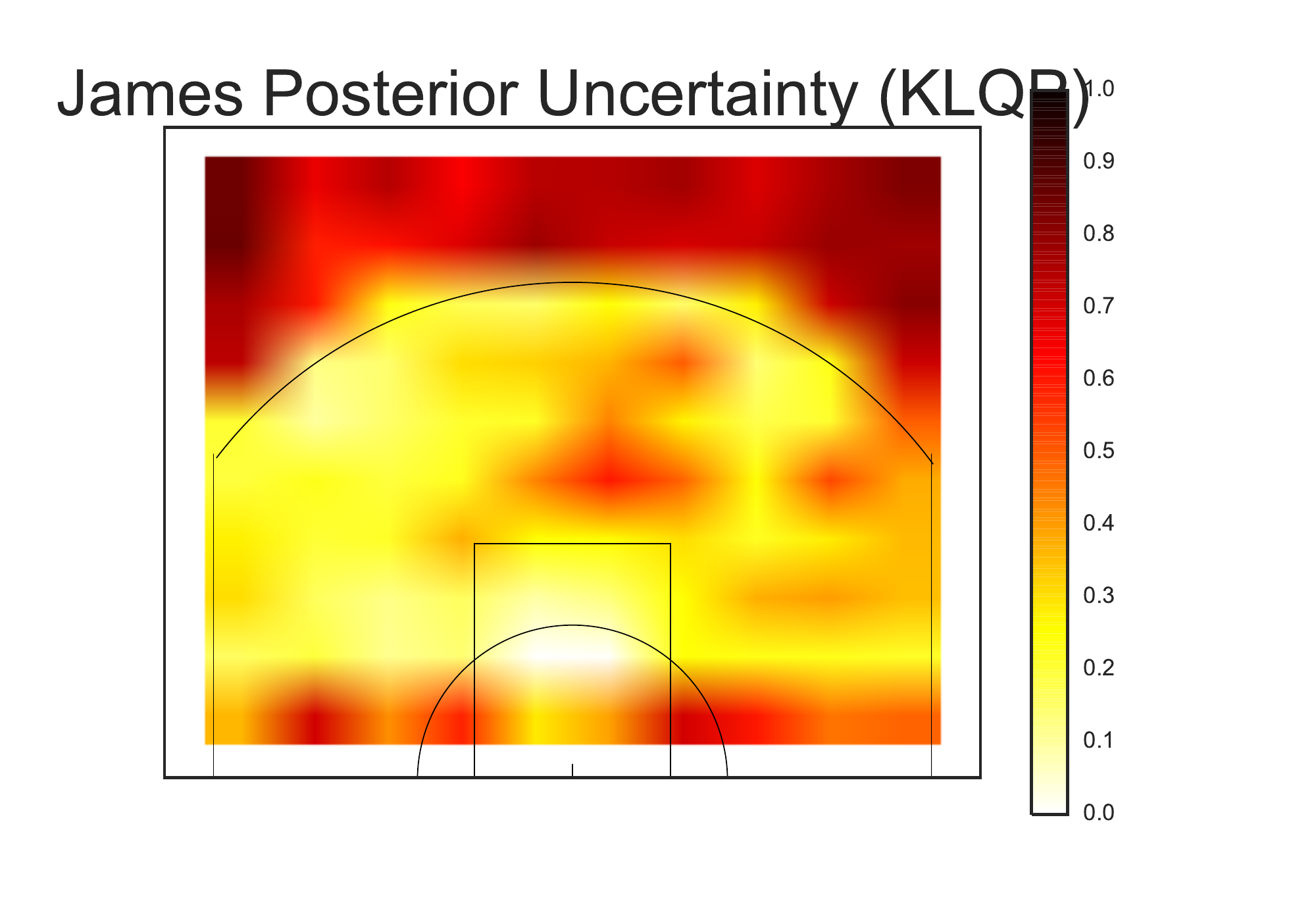}}
     {\includegraphics[scale=0.22]{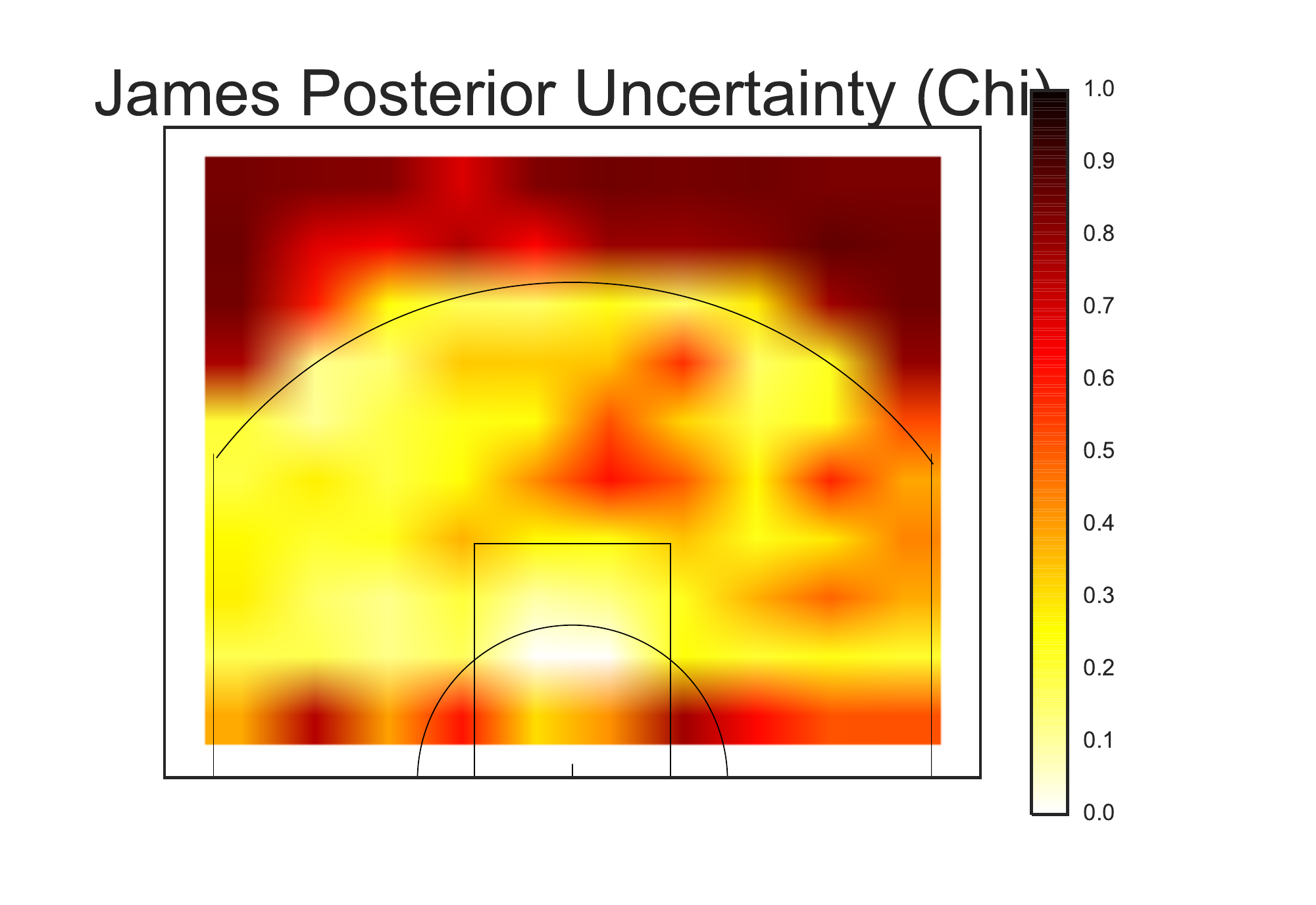}}
      {\includegraphics[scale=0.22]{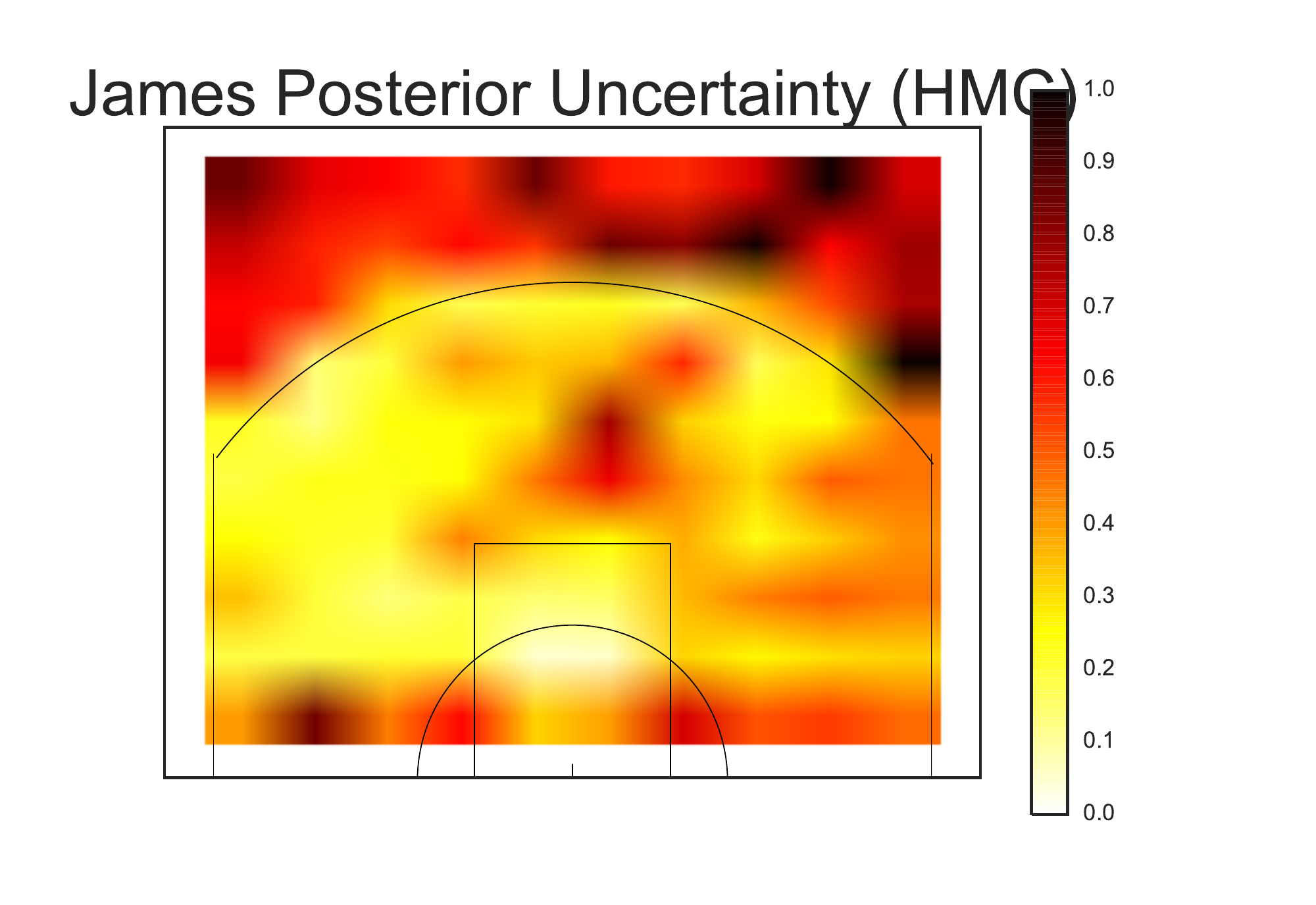}}\\
        {\includegraphics[scale=0.22]{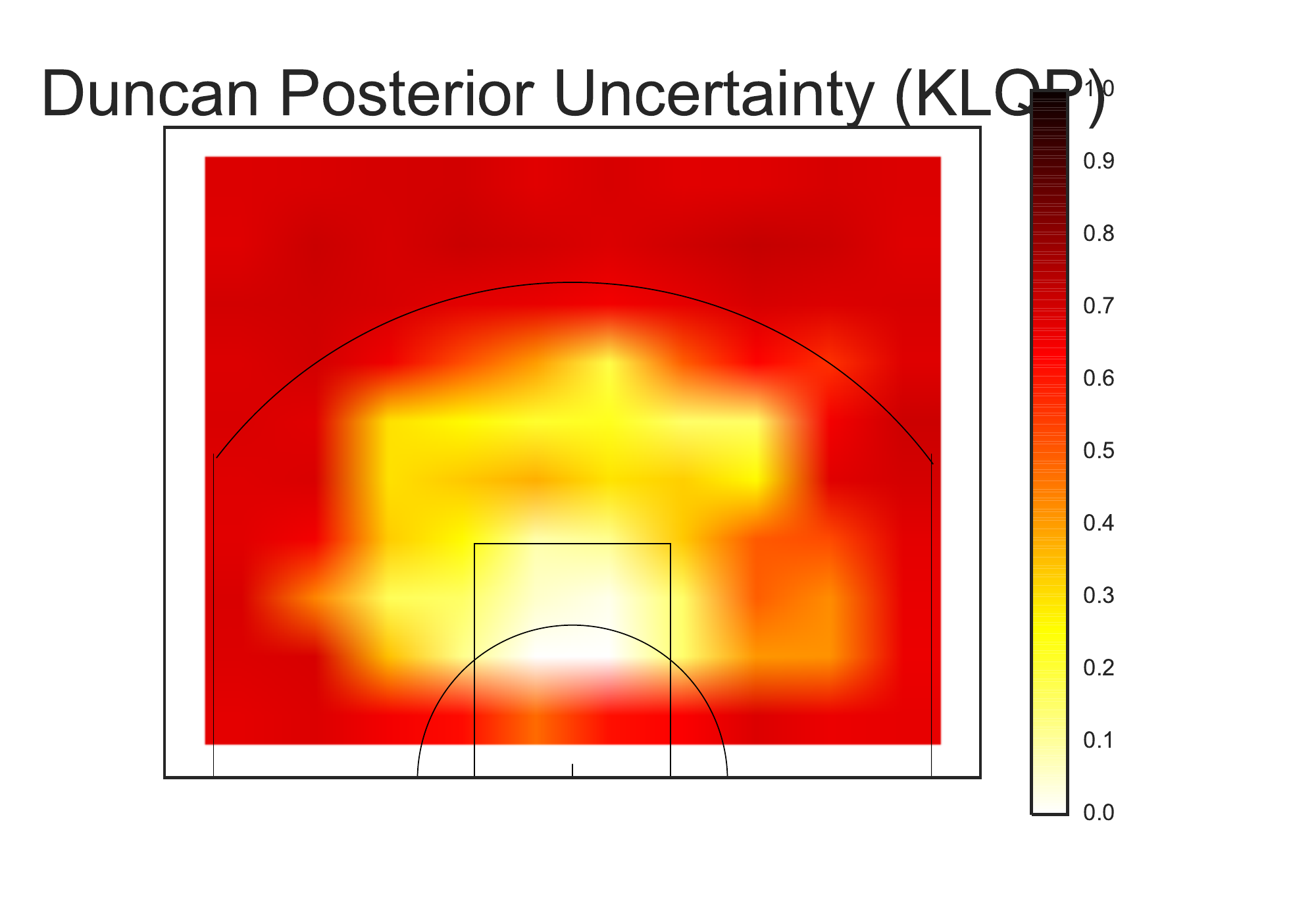}}
    {\includegraphics[scale=0.22]{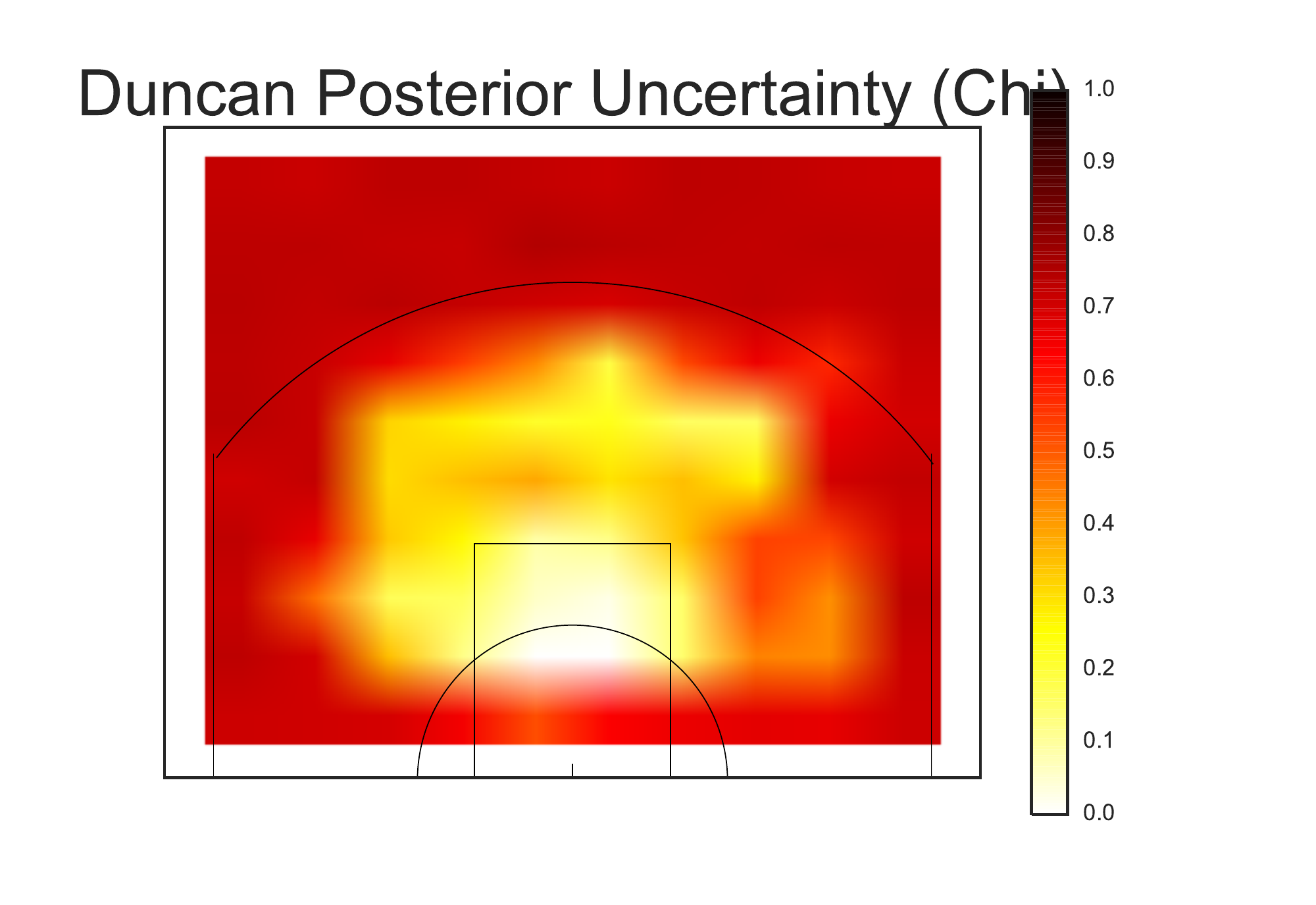}}
    {\includegraphics[scale=0.22]{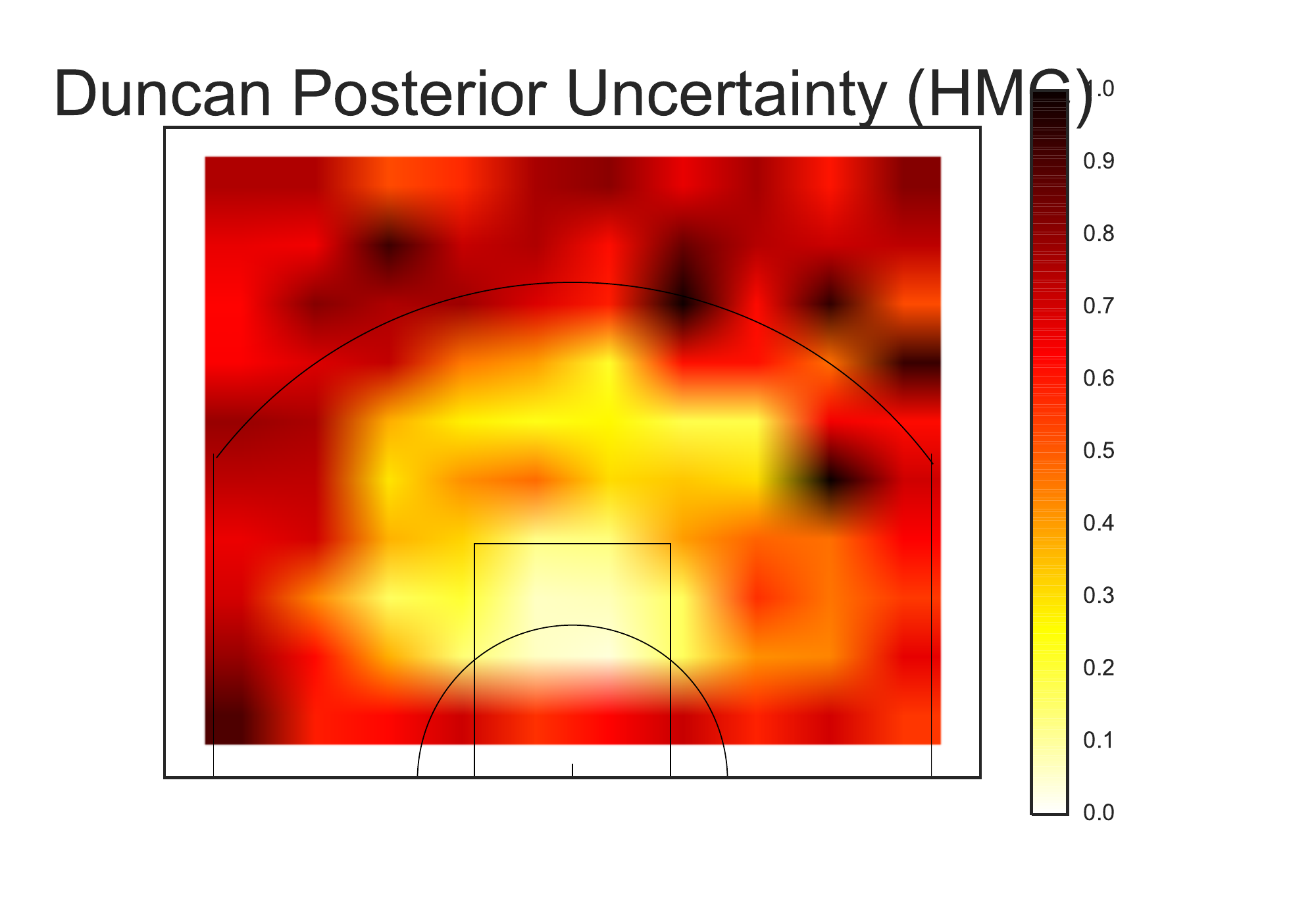}}
    \caption{More player profiles. Basketball players shooting profiles as inferred by
    \gls{BBVI} ~\citep{ranganath2014black}, \gls{chiVI} (this paper) and \gls{HMC}.
    The top row displays the raw data, consisting
    of made shots (\green{green}) and missed shots (\red{red}). The
    second and third rows display the posterior intensities inferred
    by \gls{BBVI}, \gls{chiVI} and \gls{HMC} for Lebron James
    and Tim Duncan respectively.  Both \gls{BBVI} and \gls{chiVI} nicely capture the
    shooting behavior of both players in terms of their posterior
    mean.The fourth and fifth rows display the posterior uncertainty
    inferred by \gls{BBVI}, \gls{chiVI} and \gls{HMC} for Lebron James
    and Tim Duncan respectively. Here \gls{chiVI} and \gls{BBVI}
    tend to get similar posterior uncertainty for Lebron James.
    \gls{chiVI} has better uncertainty for Tim Duncan. }
    \label{fig:players_bis}
\end{figure*}
 
\end{document}